\documentclass[10pt,twocolumn,letterpaper]{article}

\usepackage{cvpr}              
\usepackage{CJKutf8}
\usepackage{bm}
\usepackage{algorithm} 
\usepackage{algorithmic} 
\usepackage{multirow}
\usepackage{booktabs}
\usepackage{CJKutf8}
\usepackage{bbding}
\usepackage{soul}
\usepackage{amsthm}
\newtheoremstyle{mystyle}
{3pt}
{3pt}
{\itshape}
{} 
{}
{} 
{.0em}
{}
\theoremstyle{mystyle}

\usepackage[pagebackref,breaklinks,colorlinks,citecolor=cvprblue]{hyperref}
\usepackage[dvipsnames]{xcolor}
\usepackage{comment}

\def\paperID{10253} 
\def\confName{CVPR}
\def\confYear{2024}

\definecolor{cvprblue}{rgb}{0.21,0.49,0.74}
\usepackage[pagebackref,breaklinks,colorlinks,citecolor=cvprblue]{hyperref}

\title{Beyond Text: Frozen Large Language Models in Visual Signal Comprehension}

\author{Lei Zhu$^{1}$\quad Fangyun Wei$^{2}$\thanks{Corresponding author.} \quad Yanye Lu$^{1}$ \\
$^{1}$Peking University \quad  $^{2}$Microsoft Research Asia \\
{\tt\small zhulei@stu.pku.edu.cn} \quad {\tt\small fawe@microsoft.com}  \quad {\tt\small yanye.lu@pku.edu.cn}
}

\begin{document}
\maketitle

\begin{abstract}
In this work, we investigate the potential of a large language model (LLM) to directly comprehend visual signals without the necessity of fine-tuning on multi-modal datasets. The foundational concept of our method views an image as a linguistic entity, and translates it to a set of discrete words derived from the LLM's vocabulary. To achieve this, we present the Vision-to-Language Tokenizer, abbreviated as V2T Tokenizer, which transforms an image into a ``foreign language'' with the combined aid of an encoder-decoder, the LLM vocabulary, and a CLIP model. With this innovative image encoding, the LLM gains the ability not only for visual comprehension but also for image denoising and restoration in an auto-regressive fashion—crucially, without any fine-tuning. We undertake rigorous experiments to validate our method, encompassing understanding tasks like image recognition, image captioning, and visual question answering, as well as image denoising tasks like inpainting, outpainting, deblurring, and shift restoration. Code and models are available at \href{https://github.com/zh460045050/V2L-Tokenizer}{https://github.com/zh460045050/V2L-Tokenizer}.
\end{abstract}

\vspace{-5mm}
\section{Introduction}
\vspace{-1mm}
Significant advancements have been achieved in the field of natural language processing (NLP) through the deployment of large language models (LLMs), such as GPT~\cite{GPT1, GPT2, GPT3, GPT4}, PaLM~\cite{PaLM, PaLM2} and LLaMA~\cite{LLaMA, LLaMA2}. In pursuit of addressing intricate challenges necessitating the combination of text and visual understanding, scholars are broadening the capacities of the off-the-shelf LLMs. This enhancement involves the incorporation of additional visual processing components that facilitate the understanding of visual content~\cite{MiniGPT4, BLIP2, AdaptorV2, LLaVA, LQAE} or the generation of images from text~\cite{ContralGPT4, Photorealistic, SurAdapter, VisualChatGPT}. Subsequently, these improved models undergo an extra re-training or fine-tuning using various multi-modal datasets to align the visual latent space with the language latent space. Nevertheless, the refinement process generally requires a substantial amount of training resources.

\begin{figure}
\centering
\includegraphics[width=0.49\textwidth]{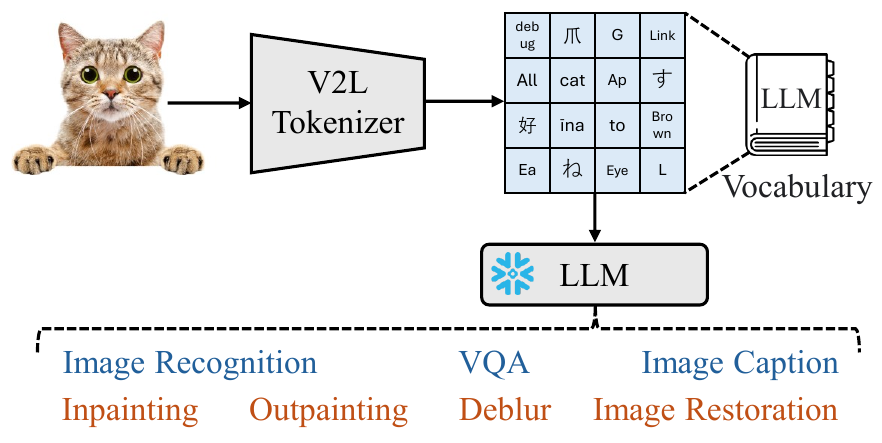}
\vspace{-7mm}
\caption{Illustration of our V2L Tokenizer (Vision-to-Language Tokenizer). The V2L Tokenizer translates an image into a collection of interpretable tokens derived from an LLM vocabulary. Subsequently, the frozen LLM can comprehend the visual signals and perform multi-modal understanding tasks (highlighted in \textcolor{RoyalBlue}{Blue}) and image denoising tasks (highlighted in \textcolor{Bittersweet}{Orange}) without the necessity of fine-tuning.}
\vspace{-4mm}
\label{fig:teaser}
\end{figure}

As illustrated in Figure~\ref{fig:teaser}, this work aims to equip a large language model with the innate ability to comprehend visual signals, importantly, without the necessity of fine-tuning. In our approach, we view each image as a linguistic entity derived from a ``foreign language'', adapting it to suit the input requirements of a plain LLM. Consequently, this alignment occurs in the input (token) space rather than in the feature space, distinguishing our work from previous multi-modal methodologies~\cite{BLIP2, MiniGPT4, Flamingo, LLaVA} that require fine-tuning for modality alignment. Thus, the fine-tuning or re-training process on multi-modal datasets is avoidable in our methodology. Our technique translates an image into a collection of discrete tokens that are within the vocabulary of the LLM. Once translated, these tokens can be fed into the LLM, enabling it to process and comprehend visual information, thereby facilitating a range of tasks involving both image understanding and denoising.

Translating an image into a set of tokens that a frozen LLM can understand is challenging. In this work, we introduce a tokenizer designed to map images (a non-linguistic modality) to the input (token) space of a frozen LLM. This tokenizer is termed the Vision-to-Language Tokenizer, or V2L Tokenizer in brief. Drawing inspiration from the triumphant advances of VQ-GAN~\cite{VQGAN}, the V2L Tokenizer employs an encoder-quantizer-decoder structure. However, its target is to translate visual information into the LLM's token space. This differs from its inspiration, which aims to learn an independent latent space solely for the purpose of image generation. Our V2L Tokenizer eschews the standard process of optimizing a randomly initialized quantizer codebook; instead, it leverages the pre-existing vocabulary of the LLM as its quantizer codebook throughout the training process. With the guidance of a quantization loss function, images are converted into a set of LLM tokens upon completion of the optimization process.

Typically, the vocabulary of an LLM consists of both full words and subword units due to the usage of language tokenizers such as BPE~\cite{BPE} and SentencePiece~\cite{SentencePiece}. Without loss of generality, the breadth of this vocabulary influences its ability to encode images into LLM tokens—a larger vocabulary usually offers more powerful representation capabilities. In our approach, we expand the LLM's vocabulary by combining its lexical items to form bigrams or trigrams, which significantly augments the representation capacity when mapping an image into the LLM tokens. In addition to converting each image patch into a language token, our V2L tokenizer includes extracting global representations for the entire image. We accomplish this by utilizing a combination of subwords, bigrams, or trigrams from the expanded LLM vocabulary to encapsulate the image's comprehensive information.

In-context learning~\cite{INCONTEXT1, INCONTEXT2, GPT3} has been shown to be highly beneficial for zero-shot inference in LLMs. This is accomplished by prefacing the instruction text with a number of domain-specific examples during the LLM inference. Our method eschews the necessity of LLM fine-tuning, instead employing in-context learning to guide the LLM in imitating the patterns presented in the given few-shot samples. This enables the model to better comprehend the ``foreign language'' (i.e., visual modality).

Experimentally, our work surpasses previous attempts~\cite{LQAE, SPAE} in this novel scenario, where an LLM is able to comprehend visual signals without any fine-tuning or re-training, encompassing understanding tasks like image captioning and visual question answering, as well as image denoising tasks like inpainting, outpainting, deblurring, and image restoration. 

\section{Related Work}

\noindent \textbf{Image Quantization:} 
The process of image quantization is designed to transform images into a series of discrete tokens derived from a codebook~\cite{VQVAE, VQVAE2, VQGAN, RQVAE, DQVAE, RegVQ, HVQVAE, ViTVQGAN, BEITv2, LMBD}. VQ-VAE~\cite{VQVAE} stands as a notable work in the field. This method employs an encoder-decoder structure to quantize images into a collection of latent, discrete codes, which are then used to reconstruct the images. VQ-GAN~\cite{VQGAN} enhances the process of codebook learning by incorporating adversarial and perceptual losses, enabling the codebook to capture more precise and finely detailed representations. Meanwhile, quantizing an image into a series of tokens enables image generation in an auto-regressive manner using GPT~\cite{GPT1, GPT2, GPT3}. RQ-VAE~\cite{RQVAE} employs a residual quantization approach, where each image patch is represented by multiple codebook tokens, to more accurately mirror the original image features. DQ-VAE~\cite{DQVAE} further present tokens of variable length to encode images, resulting in more precise and efficient tokenization. Reg-VQ~\cite{RegVQ} aims to improve the utilization of the codebook and prevent its collapse by leveraging prior distribution regularization.

\noindent  \textbf{Large Language Models.}
Large language models (LLMs)~\cite{BERT, T5, OPT, GPT3, PaLM2, LLaMA2}, especially those employing a Transformer-decoder architecture~\cite{GPT3, PaLM2, OPT, LLaMA2}, have made considerable progress in the domain of natural language processing. The process of developing an effective Large Language Model (LLM) generally unfolds in multiple phases, including initial pre-training~\cite{GPT3, PaLM, PaLM2, Chinchilla, Alexatm}, subsequent supervised fine-tuning~\cite{InstructGPT, Vincuna, Adaptor, AdaptorV2}, the training of reward models~\cite{RL, RAFT, DPO}, and the application of reinforcement learning using human feedback (RLHF)~\cite{WebGPT, LLaMA2, InstructGPT, Sparrow, GPT4} to achieve alignment with instructions. The LLaMA~\cite{LLaMA} family has been at the forefront of offering open-source LLMs, providing both aligned and non-aligned versions in an array of scales~\cite{LLaMA, LLaMA2, GLM, Alpaca, PANDA, MedGPTzh, OPT}. For instance, the LLaMA 2~\cite{LLaMA2} presents models in the sizes of 7B, 13B, and 70B parameters.

\noindent \textbf{Visual Signal Comprehension with LLMs:} 
Despite the inherent capability for natural language understanding, LLMs can also act as decoders in various vision-language applications by employing a modality bridge module to align the visual with language features~\cite{Flamingo, BLIP2, MiniGPT4, LLaVA,  LLaVAMed, AdaptorV2, LaVIN, mPLUGowl, Otter, VisonLLM, PMCVQA}. For example, Flamingo~\cite{Flamingo} utilizes billions of image-text pairs to train gated cross-attention layers that facilitate the synchronization between a frozen vision encoder and a frozen LLM. In a similar vein, BLIP-2~\cite{BLIP2} bridges the modality gap by introducing a lightweight Q-Former. This Q-Former is trained in two respective stages: one for representative learning and the other for generative learning. In addition, both MiniGPT-4~\cite{MiniGPT4} and LLaVA~\cite{LLaVA} confirm that tuning a single linear layer on high-quality instruction data, is sufficient for feature alignment. While these methods yield satisfactory results for multi-modal understanding tasks, they lack the ability to generate visual content and necessitate the collection of additional image-text pairs to train the vision-language alignment modules.  

\begin{figure*}[t]
	\centering		\includegraphics[width=0.99\textwidth]{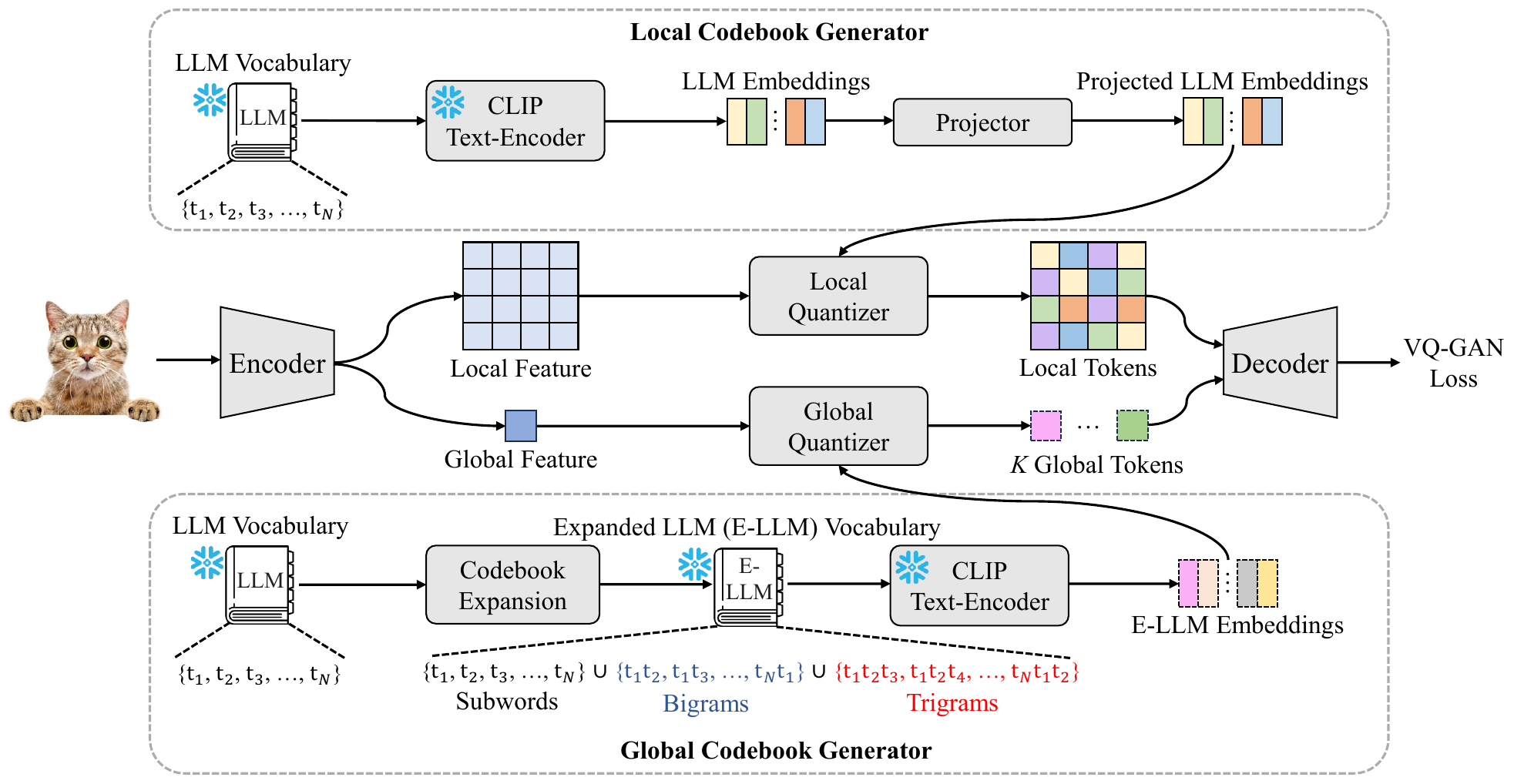}
 \vspace{-3mm}
	\caption{Overview of our Vision-to-Language Tokenizer (V2L Tokenizer). Figure~\ref{fig:intro2} illustrates its integration with a frozen LLM.}
 \vspace{-4mm}
	\label{fig:intro1}
\end{figure*}

Instead of performing multi-modal alignment in the feature space, several methods map images to the token (input) space of the LLMs by viewing images as ``foreign languages''~\cite{FROZEN, LQAE, SPAE}. For instance, LQAE~\cite{LQAE} trains a VQ-VAE tokenizer with a frozen LLM codebook to quantize an image into a set of language tokens. To enable an LLM to perform both image understanding and generation tasks, SPAE~\cite{SPAE} further enhances the quality of quantized image tokens derived from a frozen LLM codebook. It does so by incorporating a hierarchical quantization technique and semantic guidance provided by CLIP~\cite{CLIP}. However, because of the substantial difference between visual features and language token embeddings, those methods struggle to assign semantic language tokens to images. This limitation hinders LLMs from fully understanding visual signals within a given context. In contrast to the aforementioned methods, our approach introduces image quantization within a shared multi-modal space, assigning semantically meaningful language tokens to a given image. Furthermore, we separate the image tokens into two categories: global tokens, which are used for image comprehension tasks, and local tokens, which are utilized for image generation tasks. This separation is accomplished through the use of two distinct types of quantizers along with two independent codebooks.

\vspace{-1mm}
\section{Method}
\vspace{-1mm}
\begin{figure*}[t]
\centering	\includegraphics[width=0.99\textwidth]{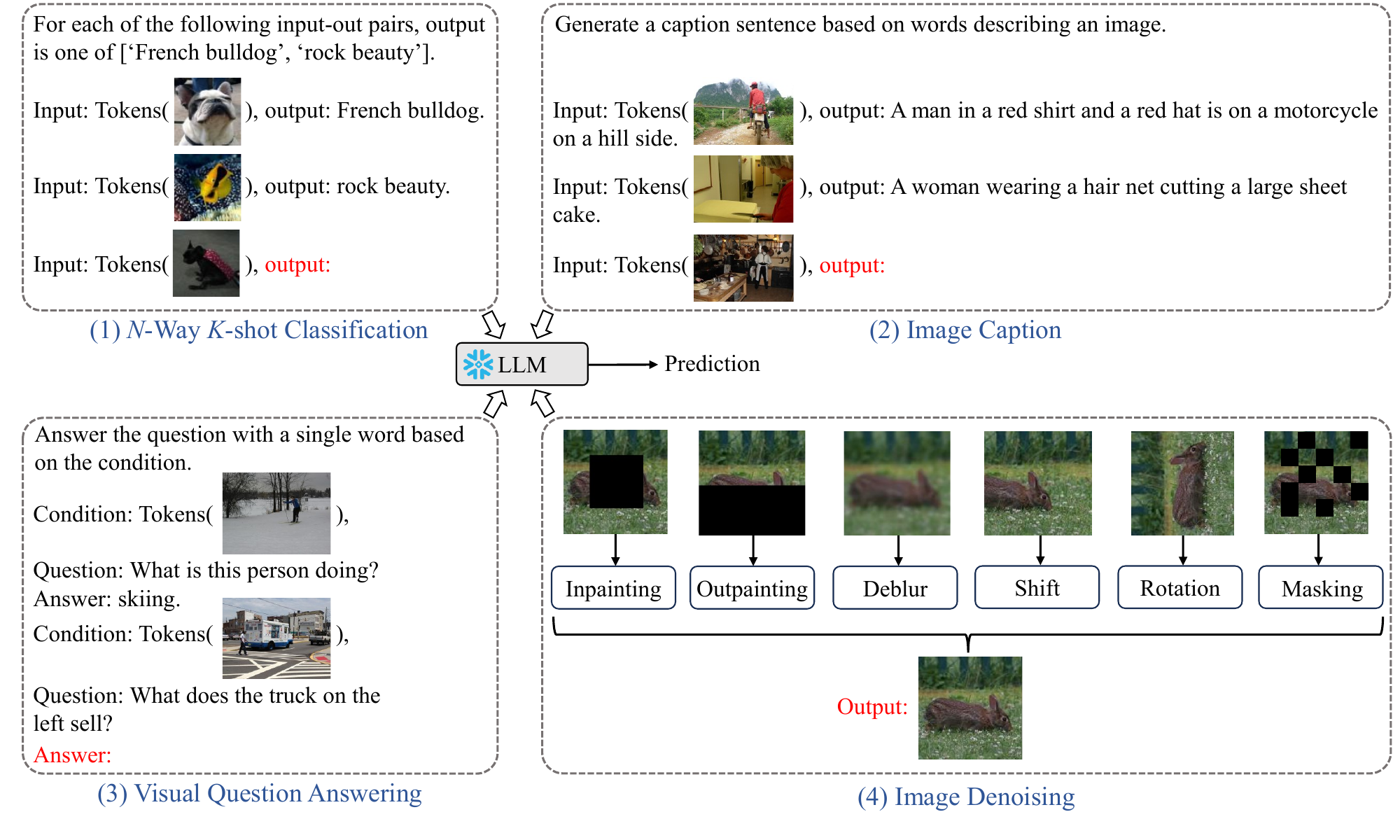}
\vspace{-3mm}
\caption{Our V2L tokenizer enables a frozen LLM to perform a series of image understanding and denoising tasks.}
\vspace{-4mm}
\label{fig:intro2}
\end{figure*}

\subsection{Problem Formulation and Overview}
\vspace{-1mm}
We view images as a ``foreign language''. Given an LLM vocabulary $\mathcal{T}=\{t_1, t_2, \dots, t_N\}$ containing $N$ language tokens, we translate an image into $K$ discrete tokens, each of which belongs to $\mathcal{T}$. This translation is accomplished by our V2L Tokenizer, as illustrated in Figure~\ref{fig:intro1}. In our implementation, an image is tokenized into $K_g$ global tokens for understanding tasks, and $K_l$ local tokens for denoising tasks, where $K = K_g + K_l$. Subsequently, as shown in Figure~\ref{fig:intro2}, we can perform a series of tasks such as image classification, image caption, visual question answering, and image denoising. This is done by feeding the concatenation of task instructions, in-context learning samples, and either global or local tokens into a frozen LLM in an auto-regressive manner.
\vspace{-1mm}
\subsection{Vision-to-Language Tokenizer}
\vspace{-1mm}
\label{sec:v2ltokenizer}
Our Vision-to-Language Tokenizer (V2L Tokenizer) adopts an encoder-quantizer-decoder structure. In total, we employ two quantizers: a local quantizer and a global quantizer. Each of these is associated with an independent, frozen codebook derived from the LLM vocabulary. An image is then quantized into $K_g$ global tokens and $K_l$ local tokens, drawn from the global and local codebooks, respectively.

\noindent\textbf{Global Codebook.} 
An LLM vocabulary comprises a set of subwords generated by language tokenizers. These subword elements, in general, tend to have limited semantic significance. To enhance the semantic representation of entities within the LLM vocabulary $\mathcal{T}$ of size $N$, we introduce a vocabulary expansion technique. This technique entails creating bigrams and trigrams by combining two or three lexical items from $\mathcal{T}$. However, it is important to note that the resulting bigrams and trigrams may not necessarily convey meaningful semantics. For instance, they may include symbols like "\#" and "!". Moreover, the generation of bigrams and trigrams leads to a vast number of possible combinations—$N^2$ bigrams and $N^3$ trigrams—which presents challenges in subsequent quantization processes.

To address this issue, we introduce a simple filter strategy. Specifically, using an image quantization dataset (such as ImageNet~\cite{IMAGENET}) and the expanded LLM vocabulary, which includes all original subwords, bigrams, and trigrams, we compute the CLIP similarities~\cite{CLIP} between each image in the dataset and every lexical item in the expanded LLM vocabulary. We then record the top-5 lexical items with the highest similarity scores for each image. Finally, we aggregate these top-5 lexical items from all images to form the final expanded LLM vocabulary, which serves as our global codebook.

\noindent\textbf{Local Codebook.}
The objective of the local codebook is to use an item from this codebook to represent a part of an image (e.g., an image patch). We use the original LLM vocabulary as our local codebook.

\noindent\textbf{Embeddings of Global and Local Codebooks.} As illustrated in Figure~\ref{fig:intro1}, we project the global codebook (i.e., the expanded LLM vocabulary) and the local codebook (i.e., the LLM vocabulary) into embeddings through the CLIP-text-encoder~\cite{CLIP}. The embeddings for the global and local codebooks are termed the LLM embeddings and the E-LLM embeddings, respectively. Additionally, we utilize a trainable projector, which is implemented as a linear layer, to further project the LLM embeddings for alignment with the visual space. The quantizers, which will be introduced later, further utilize the projected LLM embeddings (P-LLM embedding) and E-LLM embeddings to encode local and global information for an input image.

\begin{figure*}[t]
\centering	\includegraphics[width=0.99\textwidth]{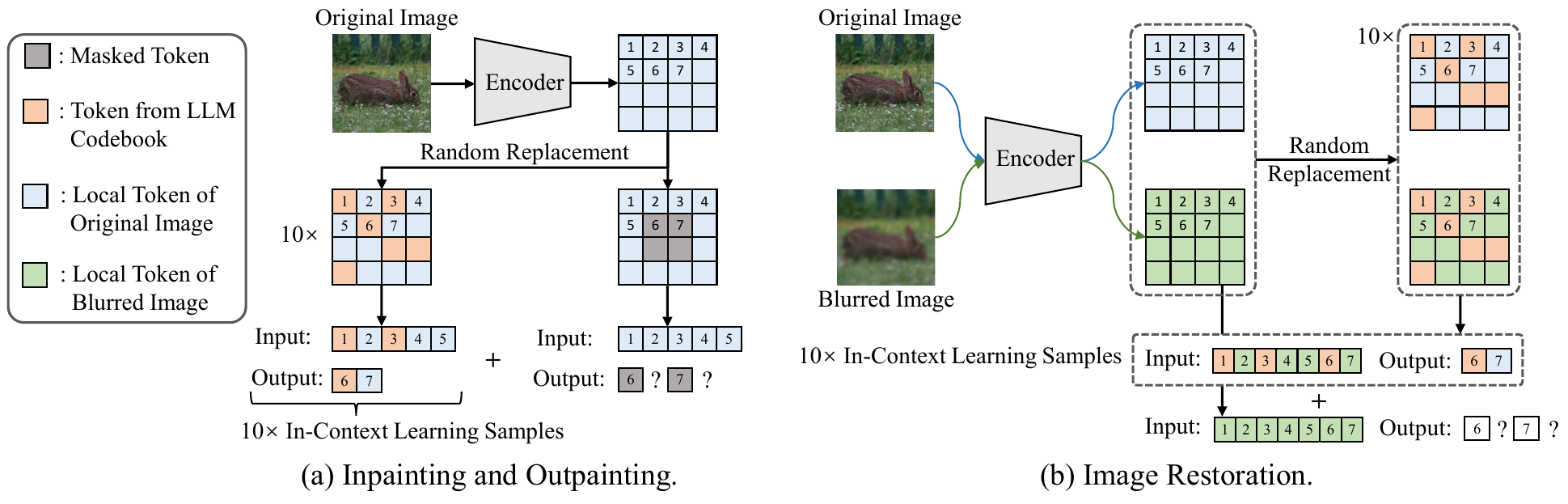}
\vspace{-3mm}
\caption{(a) We use inpainting as an example. Given an image, we first extract its local tokens $\mathcal{T}_l$. Following SPAE~\cite{SPAE}, we generate 10 copies for $\mathcal{T}_l$, termly $\{\mathcal{T}^s_l\}_{s=1}^{10}$. Each copy is a variation of $\mathcal{T}_l$ with tokens randomly replaced by those from the LLM codebook. The replacement ratios are set as [23\%, 50\%; 3\%], where 3\% denotes the incremental step. Next, an $8 \times 8$ mask (inpainting) or an $8 \times 16$ mask (outpainting) is applied to the center (inpainting) or the bottom (outpainting) of $\mathcal{T}_l$. The objective is to predict $m$ masked tokens at a time using the first $n$ tokens preceding them. The prompt is structured as follows: [\textit{Learn a new language and predict $m$ tokens following the examples. \{Input: $\mathcal{T}^s_l[n]$, output: $\mathcal{T}^s_l[m]$.\}$_{s=1}^{10}$. Input: $\mathcal{T}_l[n]$, output:}]. This prompt is then fed into the LLM, which sequentially predicts $m$ tokens. Repeating this process enables us to predict all masked tokens. Finally, we organize these predictions along with the unmasked tokens and feed the complete token map into the decoder for image restoration. 
(b) We use deblurring as an example. Both shift and rotation restorations share similar principles. The prompt is structured as follows: [\textit{Learn a new language and predict $m$ tokens following the examples. \{Input: $\overline{\mathcal{T}}^s_l[n+m]$, output: $\mathcal{T}^s_l[m]$.\}$_{s=1}^{10}$. Input: $\mathcal{T}_l[n+m]$, output:}]. In this prompt, $\overline{\mathcal{T}}_l$ denotes the local tokens of the blurred image, $\overline{\mathcal{T}}^s_l$ indicates  a variation of $\overline{\mathcal{T}}_l$ with tokens randomly replaced by those from the LLM codebook, and $\mathcal{T}^s_l$ represents the tokens of the original image, which undergo the same token replacement as $\overline{\mathcal{T}}_l^s$. By default, we set $n=16$ and $m=2$. }
\vspace{-4mm}
\label{fig:restoration_overview}
\end{figure*}

\noindent\textbf{Encoder.} Our encoder is composed of a trainable CNN encoder and a frozen CLIP-vision-encoder. The CNN encoder is identical to the one used by VQ-GAN~\cite{VQGAN}, but with modifications to the downsampling rate. We downsample the input image by a factor of 8. The CNN encoder aims to extract local information, while the CLIP-vision-encoder focuses on encoding global information. Refer to the supplementary materials for the details of the encoder. 

\noindent\textbf{Quantizers.} We use $\boldsymbol{F}\in\mathbb{R}^{h \times w \times d_l}$ to denote the feature map encoded by the CNN encoder, where $(h,w)$ is the spatial size. Similarly, $\boldsymbol{f} \in \mathbb{R}^{d_g}$ denotes the global feature encoded by the CLIP-vision-encoder, with $d_g$ representing the dimension of $\boldsymbol{f}$. Let $\mathcal{E}_{l}$ denote the set of P-LLM embeddings of the LLM vocabulary $\mathcal{T}$, and $\mathcal{E}_{g}$ represent the set of E-LLM embeddings of the expanded LLM vocabulary $\mathcal{T}_E$, respectively.

As shown in Figure~\ref{fig:intro1}, the local quantizer operates by identifying the closest embedding in $\mathcal{E}_{l}$ for each element $\boldsymbol{F}_{(i,j)} \in \mathbb{R}^{d_l}$ within $\boldsymbol{F}$, where $(i,j)$ specifies the spatial location ($1\leq i \leq h$ and $1\leq j \leq w$). The identification is based on Euclidean distance. This process yields a tokenized map $\widehat{\boldsymbol{F}}$ with the same size of $\boldsymbol{F}$. Each element $\widehat{\boldsymbol{F}}_{(i,j)} \in \mathcal{E}_l$ in $\widehat{\boldsymbol{F}}$ represents a P-LLM embedding associated with a language token belonging to $\mathcal{T}$. In total, there are $K_l = hw$ local tokens.

Similarly, the global quantizer functions by identifying the $K_g$ closest embeddings in $\mathcal{E}_{g}$ for the global feature $\boldsymbol{f}$, based on their Euclidean distance. After quantization, $\boldsymbol{f}$ is represented by the $K_g$ E-LLM embeddings $\widehat{\boldsymbol{f}} = \{\boldsymbol{e}_k \in \mathcal{E}_g\}_{k=1}^{K_g}$ associated with the corresponding language tokens $\{\boldsymbol{t}_k \in \mathcal{T}_E\}_{k=1}^{K_g}$. It should be noted that during the training of quantizers, both the LLM embeddings and the E-LLM embeddings remain frozen, as illustrated in Figure~\ref{fig:intro1}.

\noindent\textbf{Decoder.}
The objective of the decoder is to reconstruct the original image by using the local embeddings $\widehat{\boldsymbol{F}}$ and the global embeddings $\widehat{\boldsymbol{f}}$ as inputs. Our decoder is built upon the one adopted by VQ-GAN~\cite{VQGAN}, which utilizes a self-attention layer and a stack of transposed convolution layers to upsample $\widehat{\boldsymbol{F}}$ along the spatial dimension. The key distinction lies in the incorporation of $\widehat{\boldsymbol{f}}$: we inject the information of $\widehat{\boldsymbol{f}}$ into the decoding process through a cross-attention layer. In our implementation, this cross-attention layer is positioned following VQ-GAN's self-attention layer, where $\widehat{\boldsymbol{F}}$ serves as queries and $\widehat{\boldsymbol{f}}$ acts as keys. This modification does not affect the structure of the original decoder adopted by VQ-GAN. Consequently, the final output of the decoder is a tensor that matches the size of the input image.

\noindent\textbf{Loss Function.} As illustrated in Figure~\ref{fig:intro1}, we optimize only the encoder, the decoder, and the projector while freezing the LLM/E-LLM embeddings, the LLM/E-LLM vocabulary and the CLIP model. Following VQ-GAN, we define the objective function as:
\begin{equation*}
    \mathcal{L} = \mathcal{L}_{VQ} + \lambda_1 \mathcal{L}_{Perceptual} + \lambda_2 \mathcal{L}_{GAN},
\end{equation*}
where $\mathcal{L}_{VQ}$, $\mathcal{L}_{Perceptual}$ and $\mathcal{L}_{GAN}$ represent vector quantization loss, perceptual loss and
GAN loss as introduced by VQ-GAN, respectively; $\lambda_1$ and $\lambda_2$ denote the weights for the respective losses. We set $\lambda_1 = 1.0$ and $\lambda_2 = 0.1$. Refer to the original VQ-GAN~\cite{VQGAN} for more details on each type of loss.

\vspace{-1mm}
\subsection{Visual Signal Comprehension}
\vspace{-1mm}
\label{sec:vsc}
We term the language tokens associated with $\widehat{\boldsymbol{f}}$ and $\widehat{\boldsymbol{F}}$ as global tokens (denoted as $\mathcal{T}_g = \{ 
 \boldsymbol{t}_k \in \mathcal{T}_E\}_{k=1}^{K_g}$) and local tokens (denoted as $\mathcal{T}_l = \{ 
 \boldsymbol{t}_k \in \mathcal{T}\}_{k=1}^{K_l}$), respectively, with the latter being after flattening. Note that $K_l = hw$, where $(h,w)$ denote the spatial size of the feature map produced by the CNN encoder. Given an image, we first feed it into our V2L Tokenizer to generate its global tokens $\mathcal{T}_g$ and local tokens $\mathcal{T}_l$. Subsequently, we can design various prompts by combining task-specific introductions, in-context learning samples, as well as either global or local tokens, and feed the prompts into \textit{a frozen LLM} to perform a series of understanding and generation tasks, as shown in Figure~\ref{fig:intro2}. We present the prompts for each task as follows.

\begin{table*}[t]
\centering
\small
\setlength{\tabcolsep}{2.5pt}
\begin{tabular}{lcr|cccccccc|cccccccc}
\toprule
~  &  & Task Induction: &  & $\checkmark$ & $\checkmark$ & $\checkmark$ & $\checkmark$ & $\checkmark$ & $\checkmark$ & ~ 
& & $\checkmark$ & $\checkmark$ & $\checkmark$ & $\checkmark$ & $\checkmark$ & $\checkmark$ 
\\
Method &  \#Tokens & N-way K-shot: & 2-1 & 2-1 & 2-3 & 2-5 & 2-1 & 2-1 & 2-1 & Avg 
& 5-1 & 5-1 & 5-3 & 5-5 & 5-1 & 5-1 & 5-1 & Avg 
\\
~ &  & \#Repetitions: 
& 0 & 0 & 0 & 0 & 1 & 3 & 5 & ~
& 0 & 0 & 0 & 0 & 1 & 3 & 5 & ~\\
\midrule
Frozen~\cite{FROZEN} & - & - & 1.7 & 33.7 & 66.0 & 66.0 & 63.0 & 65.0 & 63.7 & 51.3 
& 0.9 & 14.5 & 34.7 & 33.8 & 33.8 & 33.3 & 32.8 & 26.3
\\
LQAE~\cite{LQAE} & 256 & GPT-3.5  & 1.5 & 35.2 & 68.2 & 69.8 & 68.5 & 68.7 & 65.9 & 54.0 
& 1.0 & 15.7 & 35.9 & 36.5 & 31.9 & 36.4 & 45.9 & 29.0\\
\midrule
SPAE~\cite{SPAE} & 5 & GPT-3.5 & 5.3 & 77.2 & 84.4 & 86.0 & 79.4 & 77.2 & 77.1 & 69.5 
& - & - & - & - & - & - & - & -\\
SPAE~\cite{SPAE} & 5 & PaLM-2 (340B) & 32.2 & 84.0 & 88.5 & 88.4 & 85.1 & 83.6 & 82.4 & 77.7 
& 23.6 & 64.2 & 68.0 & 69.9 & 63.4 & 62.0 & 60.2 & 58.8 \\
Ours & 5 & LLaMA-2 (7B) & 34.2  & 73.1  & 89.0  & 93.4  & 79.6  & 80.6  & 79.1  & 75.6  
& 36.2  & 54.6  & 88.6  & 91.1  & 70.7  & 72.8  & 74.4  & 69.8 \\
Ours & 5 & LLaMA-2 (13B) & 44.4  & 77.9  & 91.9  & 94.4  & 81.5  & 82.8  & 82.0  & 79.3 
& \textbf{45.4}  & 69.6  & 89.9  & 91.3  & 75.8  & 75.7  & 77.2  & 75.0 \\
Ours & 5 & LLaMA-2 (70B) & 41.7 & 87.1 & 94.8 & 96.1 & 88.9 & 89.2 & 89.1 & 83.9 
& \textbf{45.4} & \textbf{81.5} & 92.3 & 93.0 & 85.7 & 86.1 & 86.3 & 81.5 \\
\midrule
SPAE~\cite{SPAE} & 21 & PaLM-2 (340B) & 27.9 & 84.8 & 92.5 & 92.6 & 84.8 & 85.2 & 85.4 & 79.0 
& 20.2 & 65.1 & 73.7 & 74.3 & 66.4 & 67.0 & 66.3 & 61.9 \\
Ours & 21 & LLaMA-2 (7B) & 36.5  & 76.3  & 91.2  & 95.3  & 84.0  & 84.4  & 83.7  & 78.8  
& 37.1  & 44.8  & 91.8  & 94.0  & 73.9  & 82.2  & 85.3  & 72.7 \\
Ours & 21 & LLaMA-2 (13B) & \textbf{48.7}  & 73.1  & 92.4  & 95.7  & 80.9  & 83.8  & 82.0  & 79.5  
& 42.1  & 62.7  & 93.0  & 94.5  & 72.8  & 79.6  & 82.0  & 75.2 \\
Ours & 21& LLaMA-2 (70B) & 46.5 & \textbf{89.1} & \textbf{96.9} & \textbf{97.8} & \textbf{91.4} & \textbf{92.7} & \textbf{92.9} & \textbf{86.7} 
& 45.0 & 79.7 & \textbf{94.9} & \textbf{95.6} & \textbf{89.3} & \textbf{90.7} & \textbf{90.2} & \textbf{83.5} \\
\bottomrule
\end{tabular}
\vspace{-3mm}
\caption{Few-shot Classification on 2-way and 5-way Mini-ImageNet benchmarks.}
\vspace{-4mm}
\label{tab:2way}
\end{table*}

\noindent\textbf{N-Way K-Shot Image Classification.}
We use a 2-way K-shot classification as an example with the target of classifying images as either ``French bulldog'' or ``Rock beauty''. The prompt is structured as follows:
[\textit{For each of the following input-output pairs, output is one of [``French bulldog'', ``Rock beauty'']. \{Samples\}. Input: $\mathcal{T}_g^{Test}$, output:}], where $\mathcal{T}_g^{Test}$ denotes the global language tokens of the test image, and ``\{Samples\}'' signifies N-way K-shot samples. Each sample follows the format ``Input: $\mathcal{T}_g$, output: $L$.'', with $\mathcal{T}_g$ and $L$ denoting the corresponding global tokens and the label (either ``French bulldog'' or ``rock beauty'') of each sample, respectively.

\noindent\textbf{Image Caption.} 
We structure the prompt as follows: [\textit{Generate a caption sentence based on words describing an image. \{Samples\}. Input: $\mathcal{T}_g^{Test}$, output:}], where ``\{Samples\}'' denotes in-context learning samples. Each sample is formatted as ``Input: $\mathcal{T}_g$, output: $C$'', with $\mathcal{T}_g$ and $C$ denoting the corresponding global tokens and the caption of each sample, respectively. The LLM takes this prompt as input and auto-regressively captions the test image with global tokens $\mathcal{T}_g^{Test}$, continuing until it encounters the token ``.''.

\noindent\textbf{Visual Question Answering.} 
The prompt for VQA is designed as follows: [\textit{Answer the question with a single word based on the condition. \{Samples\}. Condition: $\mathcal{T}_g^{Test}$. Question: $Q$. Answer:}], where $\mathcal{T}_g^{Test}$ denotes the global tokens of the test image, $Q$ is the intended question, and ``\{Samples\}'' indicates in-context learning samples. Each sample has a format of ``Condition: $\mathcal{T}_g$. Question: $Q$. Answer: $A$'', with the triplet $(\mathcal{T}_g, Q, A)$ denoting the global tokens of one sample, the question related to this sample, and the ground-truth answer.

\noindent\textbf{Image Denoising.} We design several image denoising tasks following SPAE~\cite{SPAE}, including inpainting, outpainting, deblurring, shift restoration and rotation restoration. The prompts for those tasks are illustrated in Figure~\ref{fig:restoration_overview}.

\vspace{-1mm}
\section{Experiments}
\vspace{-1mm}

\begin{figure}[t]
\centering
\includegraphics[width=0.49\textwidth]{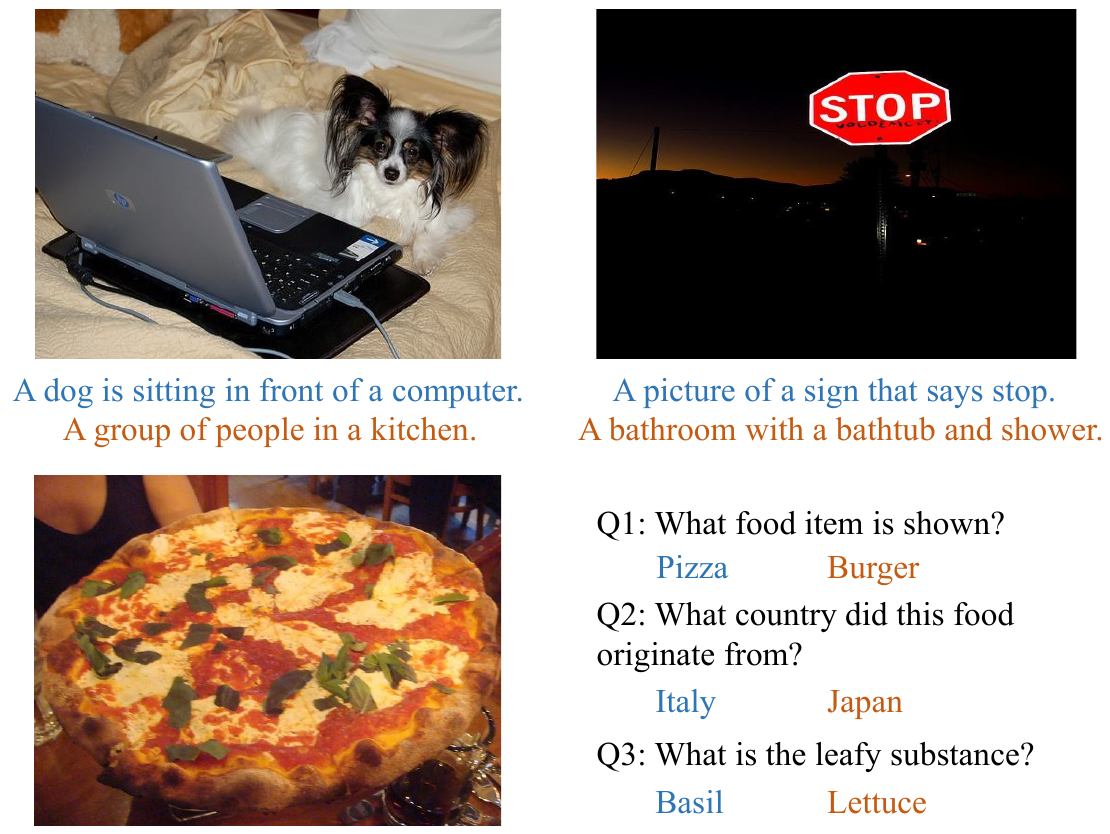}
\vspace{-7mm}
\caption{Visualizations for image caption (first row) and visual question answering (second row). \textcolor{RoyalBlue}{Blue}: ours. \textcolor{Bittersweet}{Orange}: SPAE~\cite{SPAE} (re-implementation).} 
\vspace{-4mm}
\label{fig:caption_VQA}
\end{figure}

\subsection{Settings}
\vspace{-1mm}
We adopt LLaMA 2~\cite{LLaMA2} as our LLM, which has three versions with parameters of 7B, 13B, and 70B. Its vocabulary size is $32,000$. Our local codebook retains the original vocabulary from LLaMa 2. The size of the global codebook is $11,908$ after vocabulary expansion and filtering. The CLIP model used is the one with a ViT-L/14 backbone. Images are resized to a resolution of $128\times128$ pixels and are then processed by our V2L Tokenizer, which encodes them into a $16\times16$ token map. The training is conducted on the ImageNet-1K dataset over 100 epochs using 32 NVIDIA V100 GPUs. We use the Adam optimizer, starting with a learning rate of 5e$^{-4}$, which undergoes a half-cycle cosine decay following a 5-epoch linear warm-up phase.

\vspace{-2mm}
\subsection{Image Comprehension}
\vspace{-1mm}
\noindent\textbf{Few-Shot Classification.} Following SPAE, we conduct image comprehension experiments on 2-way and 5-way Mini-ImageNet benchmarks. All few-shot samples and test images are tokenized by our V2L Tokenizer into $K_g$ global tokens. Then we structure the prompt as detailed in Section~\ref{sec:vsc} and illustrated in Figure~\ref{fig:intro2}. This prompt is then input into the LLM for the purpose of predicting the category of the test image. It is important to note that the predictions are presented in text form. A prediction is considered correct only if all the generated tokens match the tokens of the actual category name. 

Table~\ref{tab:2way} shows the comparison between our approach employing different LLaMa 2 model configurations, and prior works including LQAE~\cite{LQAE}, SPAE~\cite{SPAE}, and a baseline using a frozen language model for multimodal few-shot learning~\cite{FROZEN}. We examine various factors that could influence N-way K-shot classification, including: (1) the value of N; (2) the value of K; (3) task induction, defined as specifying the particular N-way categories in the prompt; (4) the frequency of repetitions for each few-shot sample. We have two main observations: (1) Our model surpasses the previously best approach, SPAE~\cite{SPAE}, across all scenarios, despite using smaller LLMs (our 13B/70B LLaMa 2 versus SPAE's 340B PaLM-2) and a more compact vocabulary (our 11,908 versus SPAE's 65,000); (2) The performance of our model improves as the number of tokens used to represent the image increases. This can be attributed to the introduction of the vocabulary expansion, which generates a larger pool of semantically relevant token candidates.

\noindent\textbf{Image Caption and Visual Question Answering.}
Following SPAE~\cite{SPAE}, we randomly select 10 image-caption pairs (or image-question-answer triplets) from COCO Caption~\cite{COCOCaption} (or VQA~\cite{VQA}) training set to form the in-context learning samples in the image caption (or VQA) prompt, as described in Section~\ref{sec:vsc}. By default, we utilize 21 global tokens to represent an image. The visualization results are presented in Figure~\ref{fig:caption_VQA}. Refer to supplementary materials for more results.

\begin{figure}
\centering
\includegraphics[width=0.48\textwidth]{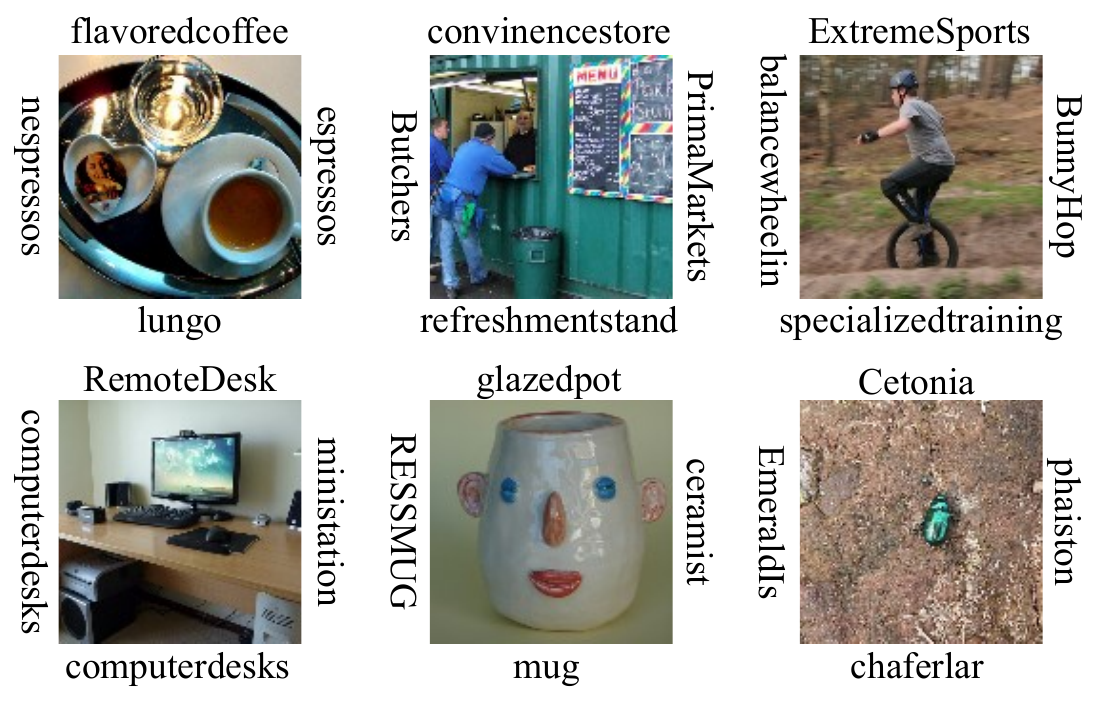}
\vspace{-8mm}
\caption{Visualization for semantic interpretation.}
\vspace{-3mm}
\label{fig:global_tokens}
\end{figure}

\begin{table}
\centering
\small
\begin{tabular}{lcccccc}
\toprule
Method & Codebook & \#Tokens & CLIP$\uparrow$ & CLIP-R$\uparrow$ \\
\midrule
SPAE~\cite{SPAE} & PaLM-2  & 5 & 0.1868 & 0.7147  \\
Ours & E-LLaMA-2  & 5 &  \textbf{0.2576} & \textbf{0.9165}  \\
\midrule
SPAE~\cite{SPAE} & PaLM-2  & 21 & 0.1815 & 0.6901 \\
Ours & E-LLaMA-2  & 21 & \textbf{0.2427} & \textbf{0.8520} \\
\bottomrule
\end{tabular}
\vspace{-3mm}
\caption{Semantic quality evaluatoin on ImageNet-1K val set. E-LLaMA-2: expanded LLaMa-2 vocabulary.}
\vspace{-4mm}
\label{tab:token_quality}
\end{table}

\begin{table}
\small
\centering	
\setlength{\tabcolsep}{1.6pt}
\begin{tabular}{lcccccc}
\toprule
Method & Codebook & \#Tokens& FID$\downarrow$ & LPIPS$\downarrow$  & PSNR$\uparrow$ \\
\midrule
VQ-GAN~\cite{VQGAN} & Learnable & 256 & 5.48 & 0.13 & - \\
VQ-GAN~\cite{VQGAN} & PaLM-2 & 256 & 7.44 & 0.17 & - \\
VQ-GAN$^{*}$~\cite{VQGAN} & LLaMA-2 & 256 & 9.51 & 0.17 & 21.48 \\
\midrule
SPAE~\cite{SPAE} & PaLM-2  & 341 & 9.49 & 0.17 & - \\
SPAE~\cite{SPAE} & PaLM-2  & 597 & 4.41 & 0.12  & - \\
SPAE~\cite{SPAE} & PaLM-2  & 1109 & 3.89 & 0.11 & - \\
\midrule
Ours & LLaMA2 & 256 & 3.41 & \textbf{0.08} & \textbf{23.56} \\
Ours & Hybrid & 277 & \textbf{2.88} & \textbf{0.08} & 23.25 \\
\bottomrule
\end{tabular}
\vspace{-3mm}
\caption{Reconstruction evaluation on ImageNet-1K val set. Hybrid: local tokens (256) and global tokens (21) are derived from the local codebook (LLaMA-2) and the global codebook (E-LLaMa-2), respectively. *: re-implementation.}
\vspace{-4mm}
\label{tab:reconstruct_eval}
\end{table}

\noindent\textbf{Semantic Interpretation.}
Figure~\ref{fig:global_tokens} visualizes the top four global tokens with the highest similarity scores for a set of six images chosen at random. Our vocabulary expansion technique effectively increases the range of semantically pertinent token options (i.e. bigrams and trigrams). Extra results are available in the supplementary materials.

In Table~\ref{tab:token_quality}, we also quantitatively evaluate the semantic quality of our global tokens, and compare the semantic quality with SPAE~\cite{SPAE} on ImageNet-1K validation set, using the CLIP score and the relative CLIP score (CLIP-R), which assess the degree of alignment between each image and its associated language tokens. We observe consistent improvements over SAPE, despite SAPE utilizing a larger vocabulary (SPAE’s 65,000 versus our 11,908).

\vspace{-1mm}
\subsection{Image Reconstruction and Denoising}
\vspace{-1mm}
\noindent\textbf{Reconstruction Evaluation.} Our V2L Tokenizer encodes an image into a set of local tokens derived from an LLM vocabulary. These encoded tokens should capture the most meaningful information, enabling the decoder to reconstruct the original image and restore any degraded (``pollutional'') images. In this study, we evaluate the reconstruction quality of our V2L Tokenizer using metrics including FID, LPIPS, and PSNR. As shown in Table~\ref{tab:reconstruct_eval}, we compare our approach with SPAE~\cite{SPAE} and VQ-GAN~\cite{VQGAN} on the ImageNet-1K validation set. In our approach, we explore two distinct setups: (1) employing the decoder from VQ-GAN without the involvement of global tokens; (2) utilizing the proposed decoder, which incorporates extra $K_g$ global tokens for the decoding process (default configuration as discussed in Section~\ref{sec:v2ltokenizer}). Our approach outperforms SPAE~\cite{SPAE} across all metrics.

\begin{table*}
\centering	
\small
\begin{tabular}{lccccccccccc}
\toprule
~ & ~ & \multicolumn{2}{c}{Inpainting} & \multicolumn{2}{c}{Outpainting} & \multicolumn{2}{c}{Deblurring} & \multicolumn{2}{c}{Rotation} & \multicolumn{2}{c}{Shift} \\
Tokenizer & LLM & FID$\downarrow$ & LPIPS$\downarrow$ & FID$\downarrow$ & LPIPS$\downarrow$ & FID$\downarrow$ & LPIPS$\downarrow$  & FID$\downarrow$ & LPIPS$\downarrow$  & FID$\downarrow$ & LPIPS$\downarrow$   \\
\midrule
VQ-GAN$^*$~\cite{VQGAN} & LLaMA-2 7B & 16.44 & 0.1404 & 18.22 & 0.1571 & 13.79 & 0.1252 & 14.08 & 0.1285 & 13.91 & 0.1270 \\
LQAE$^*$~\cite{LQAE} & LLaMA-2 7B & 18.77 & 0.1736 & 19.61 & 0.1833 & 18.09 & 0.1711 & 18.18 & 0.1725 & 18.26 & 0.1722 \\
SPAE$^*$~\cite{SPAE} & LLaMA-2 7B & 14.89 &  \textbf{0.1211} & 16.10 & \textbf{0.1363} & 15.89 & 0.1299 & 16.25 & 0.1318 & 16.55 & 0.1333 \\
Ours & LLaMA-2 7B & \textbf{13.13} & 0.1219 & \textbf{15.28} & 0.1442 & \textbf{10.09} & \textbf{0.1033} & \textbf{10.64} & \textbf{0.1064} & \textbf{10.53} & \textbf{0.1058} \\
\midrule
VQ-GAN$^*$~\cite{VQGAN} & LLaMA-2 13B & 15.56 & 0.1350 & 16.47 & 0.1449 & 14.78 & 0.1334 & 16.15 & 0.1417 & 15.60 & 0.1378 \\
LQAE$^*$~\cite{LQAE} & LLaMA-2 13B & 18.45 & 0.1720 & 18.78 & 0.1762 & 18.62 & 0.1740 & 19.04 & 0.1778 & 18.87 & 0.1770 \\
SPAE$^*$~\cite{SPAE} & LLaMA-2 13B & 13.89 & 0.1168 & 14.69 & \textbf{0.1257} & 16.46 & 0.1345 & 18.34 & 0.1436 & 17.71 & 0.1405 \\
Ours & LLaMA-2 13B & \textbf{11.70} & \textbf{0.1134} & \textbf{12.56} & 0.1275 & \textbf{10.60} & \textbf{0.1085} & \textbf{11.36} & \textbf{0.1128} & \textbf{11.84} & \textbf{0.1176} \\
\midrule
VQ-GAN$^*$~\cite{VQGAN} & LLaMA-2 70B & 14.08 & 0.1256 & 14.70 & 0.1358 & 14.30 & 0.1312 & 14.39 & 0.1313 & 14.35 & 0.1310 \\
LQAE$^*$~\cite{LQAE} & LLaMA-2 70B & 18.01 & 0.1692 & 18.54 & 0.1755 & 18.17 & 0.1713 & 18.16 & 0.1715 & 18.09 & 0.1713 \\
SPAE$^*$~\cite{SPAE} & LLaMA-2 70B & 12.79 & 0.1103 & 13.41 & 0.1191 & 18.08 & 0.1615 & 18.30 & 0.1619 & 18.19 & 0.1609 \\
Ours & LLaMA-2 70B & \textbf{10.11} & \textbf{0.1021} & \textbf{10.73} & \textbf{0.1128} & \textbf{10.42} & \textbf{0.1058} & \textbf{10.48} & \textbf{0.1058} & \textbf{10.79} & \textbf{0.1093} \\
\bottomrule
\end{tabular}
\vspace{-3mm}
\caption{Quantitative evaluation across five denoising restoration tasks. *: re-implementation. }
\vspace{-3mm}
\label{tab:denoise_generation}
\end{table*}

\begin{figure*}
\centering
\includegraphics[width=1.00\textwidth]{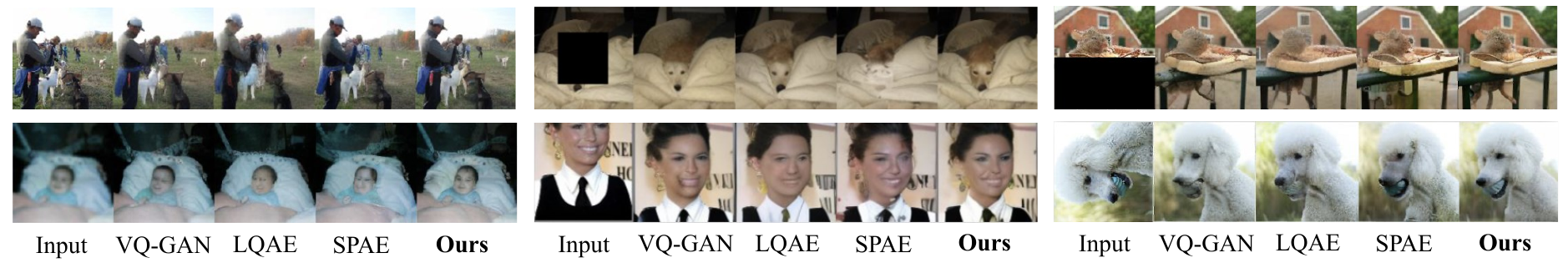}
\vspace{-8mm}
\caption{From left-to-right, top-to-bottom: visualizations for image reconstruction, inpainting, outpainting, deblurring, shift restoration and rotation restoration. We re-implement VQ-GAN~\cite{VQGAN}, LQAE~\cite{LQAE} and SPAE~\cite{SPAE} using a vocabulary size of 32,000 and 256 local tokens for a fair comparison.}
\vspace{-5mm}
\label{fig:restoration_results}
\end{figure*}

\noindent\textbf{Image Denoising.} 
We introduce the prompts used for inpainting, outpainting, deblurring, shift and rotation restorations, along with the process of restoring polluted images, as shown in Figure~\ref{fig:restoration_overview}. In Table~\ref{tab:denoise_generation}, we study two factors impacting the quality of these five in-context image denoising tasks: (1) the image tokenizer, which encodes an image into a set of tokens; (2) the LLM, which aims to predict the local tokens of the original images given the tokens of the polluted images, with the aid of in-context learning samples encoded by the tokenizer. The tokenizers used for comparison include VQ-GAN~\cite{VQGAN}, LQAE~\cite{LQAE}, and SPAE~\cite{SPAE}. We randomly select 5,000 images from the ImageNet-1K validation set to form our evaluation set. We use FID and LPIPS scores as metrics. Our V2L Tokenizer outperforms others across the five tasks on almost all metrics. This achievement is attributed to the alignment of image features with the token space of the frozen LLM. We also show several qualitative results in Figure~\ref{fig:restoration_results}. More visualizations can be found in the supplementary materials.

\begin{figure}
\centering
\includegraphics[width=0.475\textwidth]{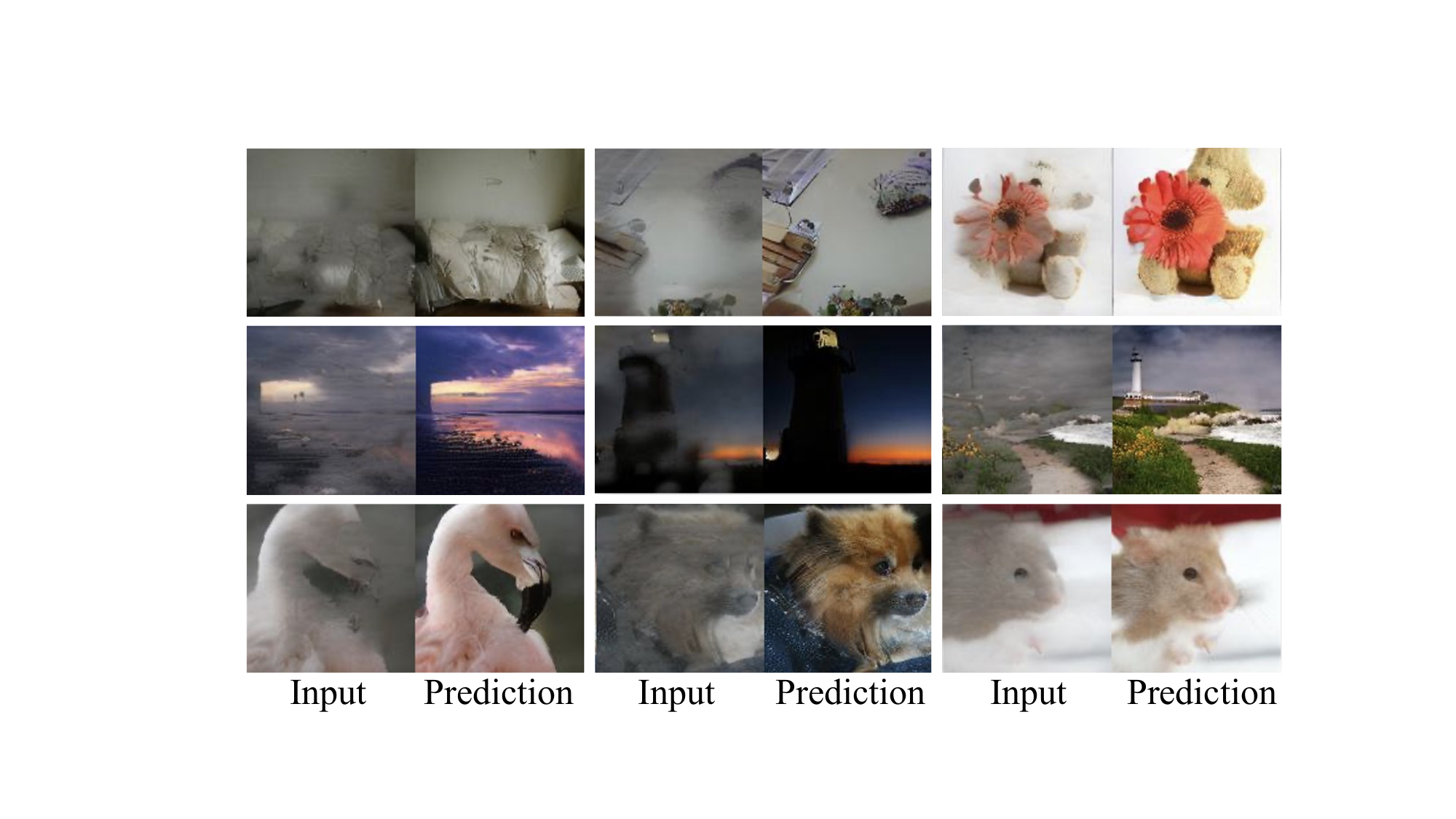}
\vspace{-8mm}
\caption{Visualizations for masked image restoration.}
\vspace{-6mm}
\label{fig:restoration_masked_images}
\end{figure}

\noindent\textbf{Masked Image Restoration.} Given an image from the ImageNet-1K validation set, we first extract its global and local tokens through our V2L Tokenizer. Subsequently, we apply random masking to 30\% of these local tokens. To predict the masked tokens, we employ a LoRA-tuned~\cite{LORA} 7B LLaMa-2 model (details on tuning are provided in the supplementary materials). The next step involves integrating the predictions for the masked tokens with the unmasked tokens, which are then input into the decoder for image reconstruction. The qualitative results of this visual signal restoration process are illustrated in Figure~\ref{fig:restoration_masked_images}. For visualization purposes, the masked images (``input'') presented are generated by combining the unmasked local tokens of the original image with the masked tokens which have been set to zero, before being processed through the decoder.

\vspace{-3mm}
\section{Conclusion}
\vspace{-2mm}
In this paper, we view images as a ``foreign language'', and introduce a V2L Tokenizer, which maps continuous visual signals to the token space of an LLM. Our method enables a frozen LLM  to understand visual signals without the necessity for resource-intensive fine-tuning on multi-modal datasets. The V2T Tokenizer processes an image by generating both global and local tokens. The global tokens are crafted to capture essential semantic information with the aid of the proposed vocabulary expansion technique. This enables the execution of tasks like image recognition, image captioning and VQA. In contrast, local tokens are designed to extract detailed, patch-level features from images, facilitating image denoising tasks such as inpainting and deblurring. Extensive quantitative and qualitative experiments validate the superiority of our approach over the prior attempts in this direction.
\clearpage
{
    \small
    \bibliographystyle{ieeenat_fullname}
    \bibliography{main}
}
\clearpage

\appendix
\section{More Implementation Details}
\begin{figure*}[t]
\centering
\includegraphics[width=0.99\textwidth]{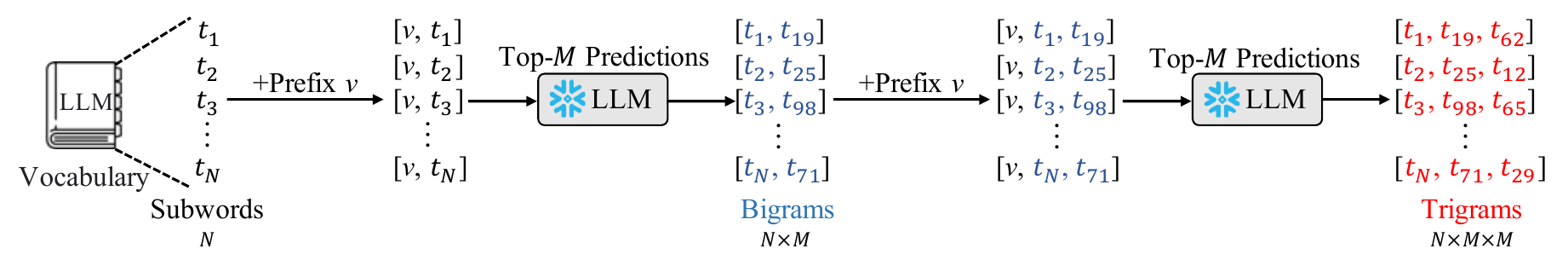}
\caption{Illustration of the vocabulary expansion strategy. In this figure, we set $M=1$ for illustrative purposes. The prefix $v$ corresponds to the text phrase ``a photo of''.}
\label{fig:global_generator}
\end{figure*}

\textbf{Global Codebook Generation.}
To generate the global codebook, we introduce a two-phase process: (1) expanding the LLM vocabulary through the proposed vocabulary expansion technique (as shown in Figure~\ref{fig:global_generator}); (2) applying a filtering strategy to further eliminate the entries with less semantic meaning.

We use $\mathcal{T}$ to represent the original LLM vocabulary and denote its size by $N$. To generate bigrams, for each $t \in \mathcal{T}$, we first input the concatenation of a text prefix (e.g., ``a photo of'') and $t$ into the LLM. The LLM predicts the next word in an auto-regressive manner. We record the top-$M$ predictions (where $M$ is 1 by default) with the highest confidences, denoted as $\{t^*_1,\dots,t^*_M\}$. The bigrams for each $t \in \mathcal{T}$ are represented by $\{[t, t^*_1], \dots, [t, t^*_M]\}$. This process is repeated for all subwords in the LLM vocabulary. Ultimately, we collect a set of bigrams, denoted as $\mathcal{T}_{Bi}$, which has a size of $N \times M$. Similarly, we can build a trigram set $\mathcal{T}_{Tri}$ by feeding each bigram in $\mathcal{T}_{Bi}$ into the LLM for next-word prediction. The resulting $\mathcal{T}_{Tri}$ has a size of $N \times M \times M$. We use $\{\mathcal{T}, \mathcal{T}_{Bi}, \mathcal{T}_{Tri}\}$ to represent the expanded LLM vocabulary.

For the filtering process, we compute the CLIP similarities between each image in the training set and every entry in the expanded LLM vocabulary $\{\mathcal{T}, \mathcal{T}_{Bi}, \mathcal{T}_{Tri}\}$. We then record the top-5 entries with the highest similarity scores for each image. Finally, we aggregate these entries from all images to form the final expanded LLM vocabulary, which serves as our global codebook $\mathcal{T}_E$.

\begin{figure*}[t]
\centering
\includegraphics[width=0.99\textwidth]{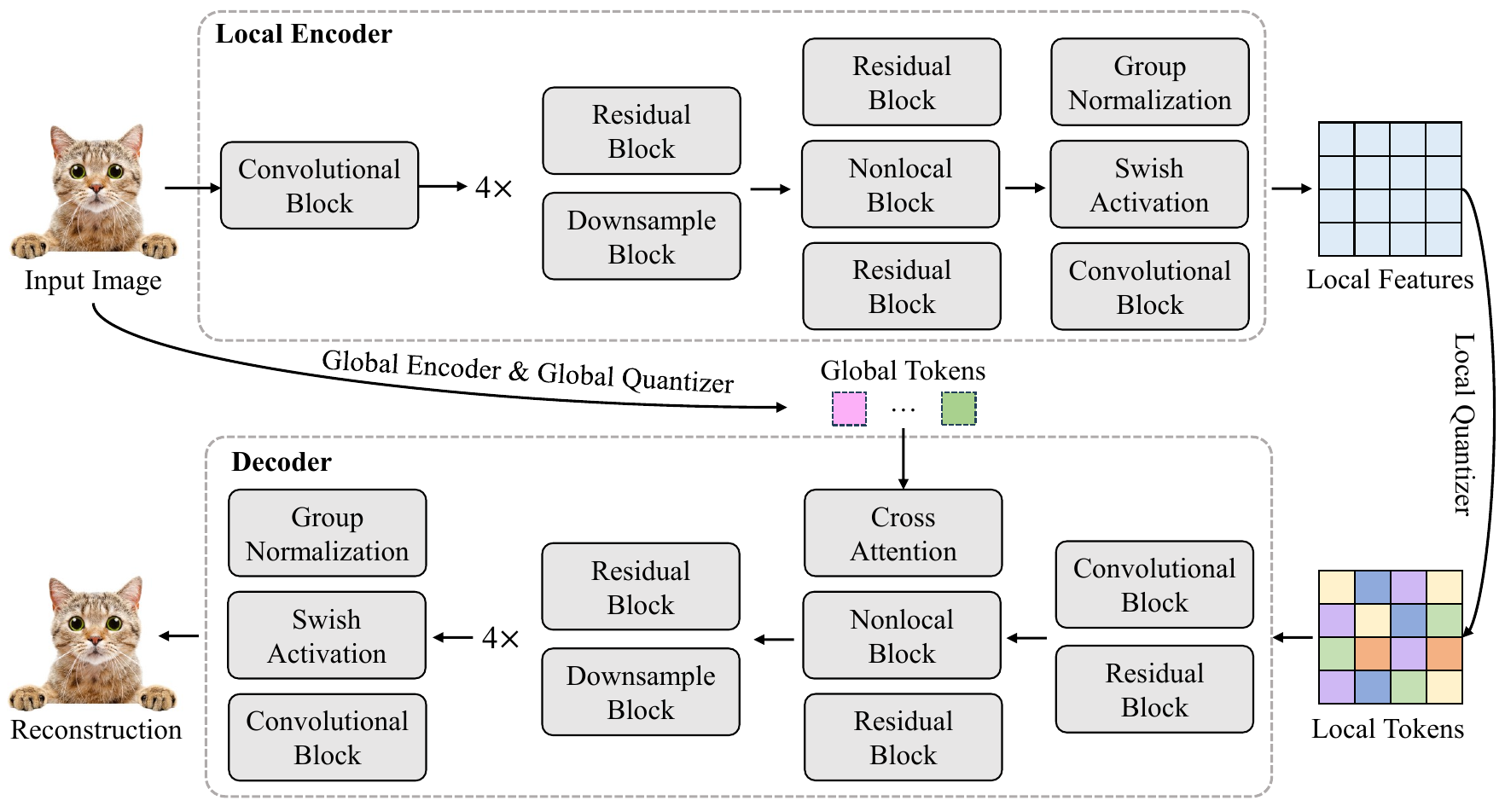}
\vspace{-2mm}
\caption{Illustration of the local encoder and the decoder of our V2L Tokenizer.}
\label{fig:structure}
\vspace{-4mm}
\end{figure*}

\noindent \textbf{Encoder and Decoder Structures.}
Figure~\ref{fig:structure} details the implementation of our V2L Tokenizer's local encoder and decoder. Specifically, the local encoder shares the same basic structure as VQ-GAN~\cite{VQGAN}, utilizing four residual blocks with channel dimensions [128, 256, 256, 512] to downsample the input image by a factor of 8. Similarly, our decoder mirrors the encoder's structure, employing four residual blocks with channel dimensions [512, 256, 256, 128] to upsample the image back to its original resolution. We integrate the information from global tokens into the decoding process through a cross-attention layer, which is added before the self-attention layer in the nonlocal block.

\noindent \textbf{Vector Quantization Loss.} 
The proposed V2L Tokenizer requires optimization of the encoder, the decoder and the projector. Thus, we follow VQ-VAE~\cite{VQVAE} and VQGAN~\cite{VQGAN} to implement our vector quantization loss, utilizing a straight-through gradient estimator for optimization:
\begin{equation*}
\mathcal{L}_{vq} = ||\bm{X} - \bm{\hat{X}}||^{2} + ||sg(\bm{F}) - \bm{\hat{F}} || + \beta || sg(\bm{\hat{F}}) - \bm{F} ||
\label{eq:vq}
\end{equation*}
where $sg(\cdot)$ denotes the stop-gradient operation. Note that our method involves a trainable projector to produce codebook embeddings. Thus, unlike LQAE~\cite{LQAE} and SPAE~\cite{SPAE}, the second term in the above equation is also necessary. We set $\beta$ to 0.3.

\noindent \textbf{Tuning LLaMA-2 with the V2L Tokenizer.} To enhance the image generation task, we propose to fine-tune an LLM model. This process begins with the V2L Tokenizer generating both global and local tokens for the training images. Subsequently, the global tokens are employed as a ``text prefix''. We then concatenate these global tokens with the local tokens and input them into the LLM. The auto-regression loss is applied only to the local tokens. Due to resource limitations, we fine-tune a 7B LLaMA-2 model using LoRA~\cite{LORA} on 12 randomly selected classes from ImageNet training dataset over 100K iterations using 32$\times$ NVIDIA V100 GPUs. LoRA weights are integrated into the query and key projection matrixes, with the hyper-parameter setting of $r=4$, $\alpha=32$. For optimization, we use Adam optimizer, starting with a learning rate of 3e$^{-4}$. This rate undergoes half-cycle cosine decay after a 5-epoch linear warm-up phase. Consequently, the tuned model is able to predict masked tokens in an auto-regressive manner. The predicted token map is input into the decoder of the V2L tokenizer to generate the reconstructed image, as demonstrated in Section 4.3 of our main paper.

\section{More Ablation Studies}

\begin{table*}
\centering	
\begin{tabular}{ccrcccccccc}
\toprule
~  &  & Task Induction: &  & $\checkmark$ & $\checkmark$ & $\checkmark$ & $\checkmark$ & $\checkmark$ & $\checkmark$ & ~ \\
Method & \#Tokens & Inner-shot: & 1 & 1 & 3 & 5 & 1 & 1 & 1 & Avg \\
~ &  & Repeats: & 0 & 0 & 0 & 0 & 1 & 3 & 5 & ~ \\
\midrule
Subword & \multirow{3}{*}{5} & \multirow{3}{*}{LLaMA-2 (70B)} & 31.8  & 65.6  & 82.8  & 85.6 & 68.8 & 69.9 & 69.3  & 67.7  \\
Bigram & ~ & ~ & 40.6  & 83.1 & 91.7 & 92.6  & 86.5  & 87.0 & 86.9 & 81.2  \\
Trigram & ~ & ~ & 41.7 & 87.1 & 94.8 & 96.1 & 88.9 & 89.2 & 89.1 & 83.9 \\
\midrule
Subword & \multirow{3}{*}{21} & \multirow{3}{*}{LLaMA-2 (70B)} & 34.3  & 74.1  & 90.1  & 91.8 & 79.6 & 80.2 & 80.7  & 75.8  \\
Bigram & ~ & ~ & 44.8  & 84.1 & 95.0 & 95.5  & 91.6  & 92.3  & 92.5 & 85.1  \\
Trigram & ~ & ~ & 46.5 & 89.1 & 96.9 & 97.8 & 91.4 & 92.7 & 92.9 & 86.7 \\
\bottomrule
\end{tabular}
\caption{Ablation study for the proposed vocabulary expansion strategy on the 5-way-K-shot Mini-ImageNet classification benchmark. }
\vspace{-2mm}
\label{tab:ablation_global}
\end{table*}

\begin{table}
\centering	
\small
\setlength{\tabcolsep}{3pt}
\begin{tabular}{cccccccc}
\toprule
Vocabulary &  Embedding & FID$\downarrow$ & LPIPS$\downarrow$ & PSNR$\uparrow$ \\
\midrule
LLaMA-2 &  LLaMA-2 & 9.51 & 0.17 & 21.48 \\
LLaMA-2 & CLIP & 4.58 & 0.11& 23.58 \\
LLaMA-2 & P-LLaMA-2 & 3.41 & 0.08 & 23.56 \\
\bottomrule
\end{tabular}
\caption{Ablation study on various LLM embeddings. We report results on ImageNet-1K val set.}
\label{tab:ablation_local}
\end{table}

\noindent \textbf{Vocabulary Expansion.}
We study the effectiveness of the proposed vocabulary expansion strategy on the 5-way-K-shot Mini-ImageNet classification benchmark. Our studies include three scenarios: utilizing the original LLM vocabulary without expansion (Subword), applying bigram expansion (Bigram), and employing trigram expansion (Trigram). The results of these scenarios are detailed in Table~\ref{tab:ablation_global}. The bigram expansion approach surpasses the non-expansion method by an average accuracy increase of +13.5 and +9.3 points with 5 and 21 global tokens, respectively. Implementing trigram expansion further elevates the average accuracy to 83.9 and 86.7. The findings demonstrate that employing vocabulary expansion significantly improves the semantic richness of the terms in the expanded LLM vocabulary, leading to enhanced classification accuracy.

\noindent \textbf{Embeddings of Local Codebook.} As shown in Figure 2 of the main paper, we introduce a trainable projector to project the LLM embeddings into a visual space, which enhances reconstruction quality.  Table~\ref{tab:ablation_local} presents our investigation of various LLM embeddings, including the default projected LLM embeddings (P-LLaMA-2), the original LLM embeddings (LLaMa-2), and those produced by the CLIP-text-encoder (CLIP). We observe that utilizing the CLIP text encoder for extracting language embeddings significantly boosts the quality of reconstruction. This improvement likely stems from the CLIP model's inherent alignment between linguistic and visual spaces. By introducing a trainable projector, this alignment is further refined, leading to superior reconstruction performance.

\begin{figure}[t]
    \vspace{-2mm}
	\centering
	\begin{subfigure}{0.4\textwidth}
		\centering
		\includegraphics[width=0.99\textwidth]{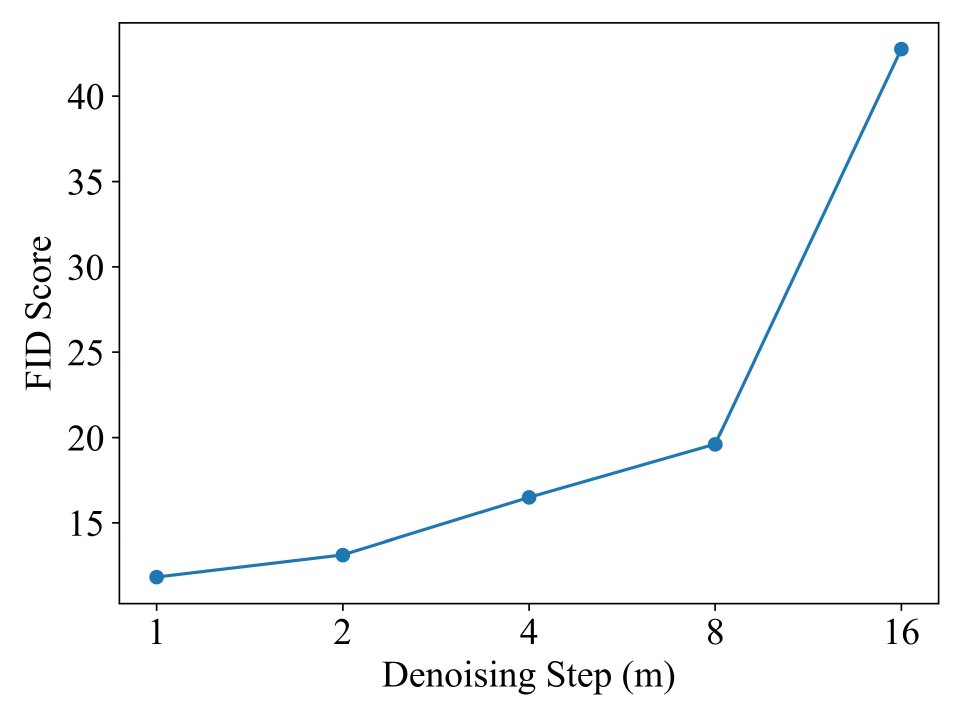}
		\caption{FID score v.s. denoising step.}
	\end{subfigure}
	\begin{subfigure}{0.4\textwidth}
		\centering
		\includegraphics[width=0.99\textwidth]{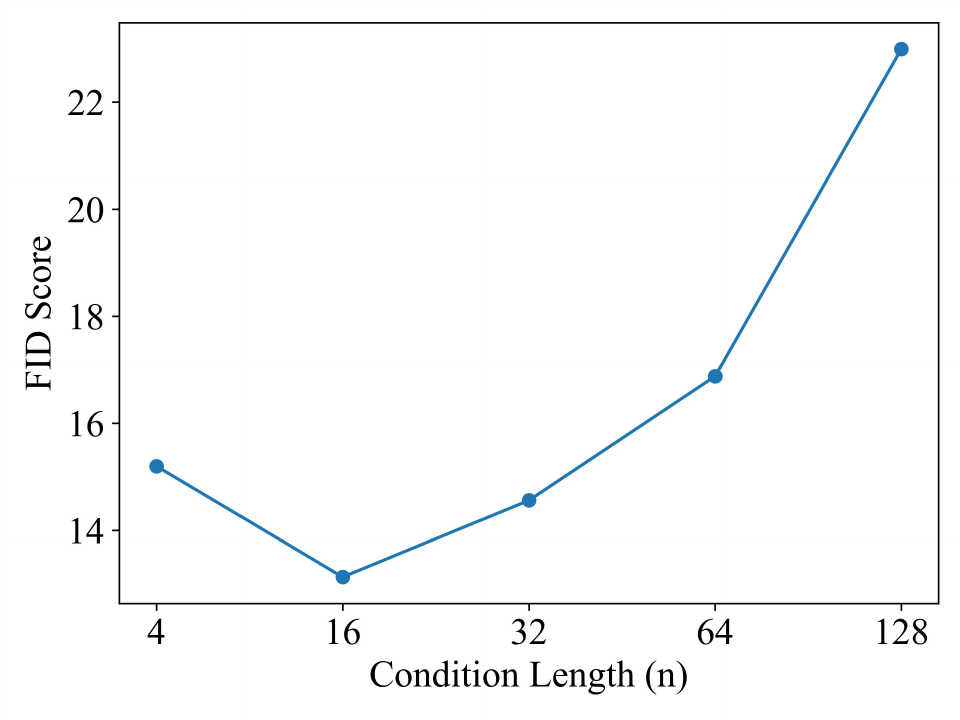}
		\caption{FID score v.s. condition length.}
	\end{subfigure}
\caption{Ablation study on the denoising step (m) and the condition length (n) for the image inpainting task, using a 7B LLaMA-2.}
\vspace{-5mm}
\label{fig:generation}
\end{figure}

\noindent \textbf{Denoising Step and Condition Length.} As shown in Figure~4 of the main paper, we denoise $m$ masked tokens at a time using $n$ tokens preceding them for the inpainting task, where $m$ and $n$ denote denoising step and condition length, respectively. We vary the values of $m$ and $n$ and report the FID scores for inpainting task in Figure~\ref{fig:generation}. As the denoising step increases, the performance decreases. 
Additionally, an excessively long condition length leads to suboptimal performance since the LLM struggles to handle the complex context of a new ``foreign language'' in the visual modality.

\section{More Qualitative Results}

\begin{figure*}[t]
\centering
\includegraphics[width=0.99\textwidth]{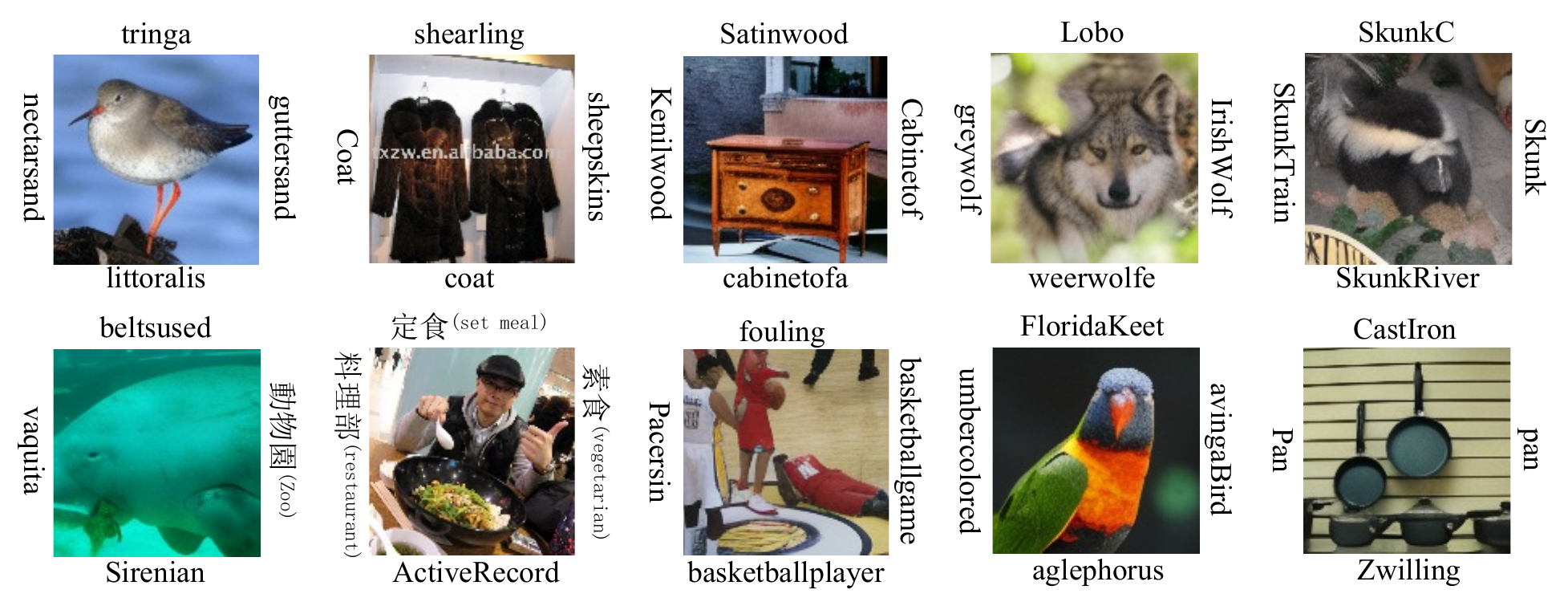}
\caption{More visualizations for semantic interpretation.}
\label{fig:global_tok}
\end{figure*}

\noindent \textbf{Semantic Interpretation.}
We provide qualitative results for semantic interpretation in Figure~6 of the main paper. Here, we show additional visualizations in Figure~\ref{fig:global_tok}.

\begin{figure*}[t]
\centering
\includegraphics[width=0.99\textwidth]{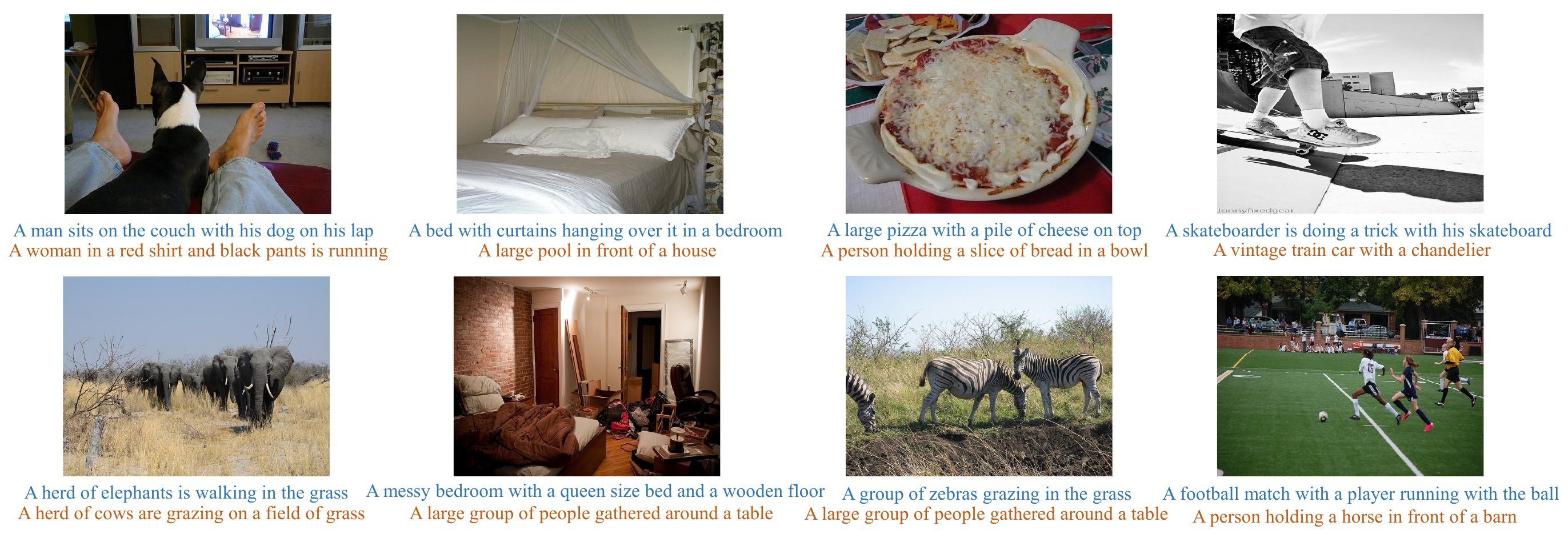}
\caption{Visualizations for image caption. \textcolor{RoyalBlue}{Blue}: ours. \textcolor{Bittersweet}{Orange}: SPAE~\cite{SPAE} (re-implementation).}
\label{fig:caption}
\end{figure*}

\begin{figure*}[t]
\centering
\includegraphics[width=0.99\textwidth]{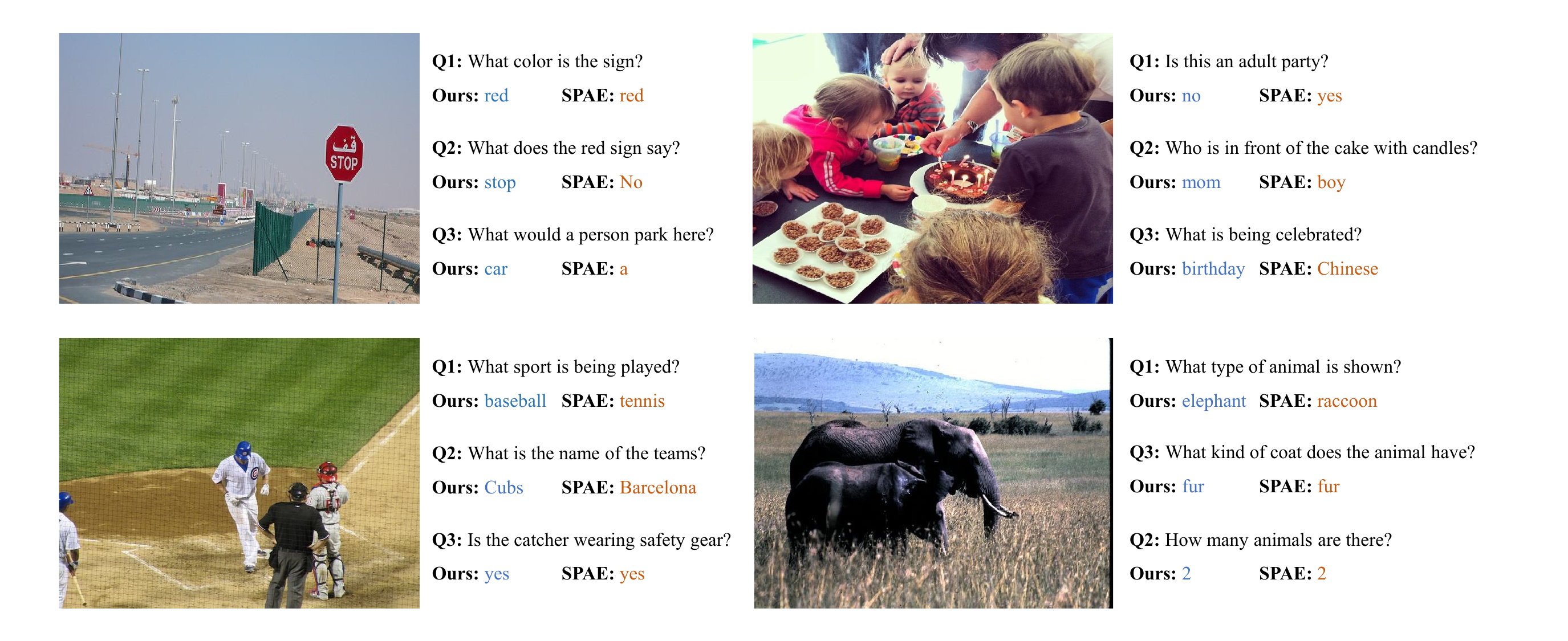}
\caption{Visualizations for visual question answering. \textcolor{RoyalBlue}{Blue}: ours. \textcolor{Bittersweet}{Orange}: SPAE~\cite{SPAE} (re-implementation).}
\label{fig:vqa}
\end{figure*}

\noindent \textbf{Image Captioning and Visual Question Answering.}
Figure~5 of the main paper visualizes the results of image captioning and VQA. In Figures~\ref{fig:caption} and~\ref{fig:vqa}, we compare our approach with SAPE~\cite{SPAE} using additional samples. Our model consistently generates more reasonable image captions and provides more accurate answers.

\begin{figure*}[t]
\centering
\includegraphics[width=0.99\textwidth]{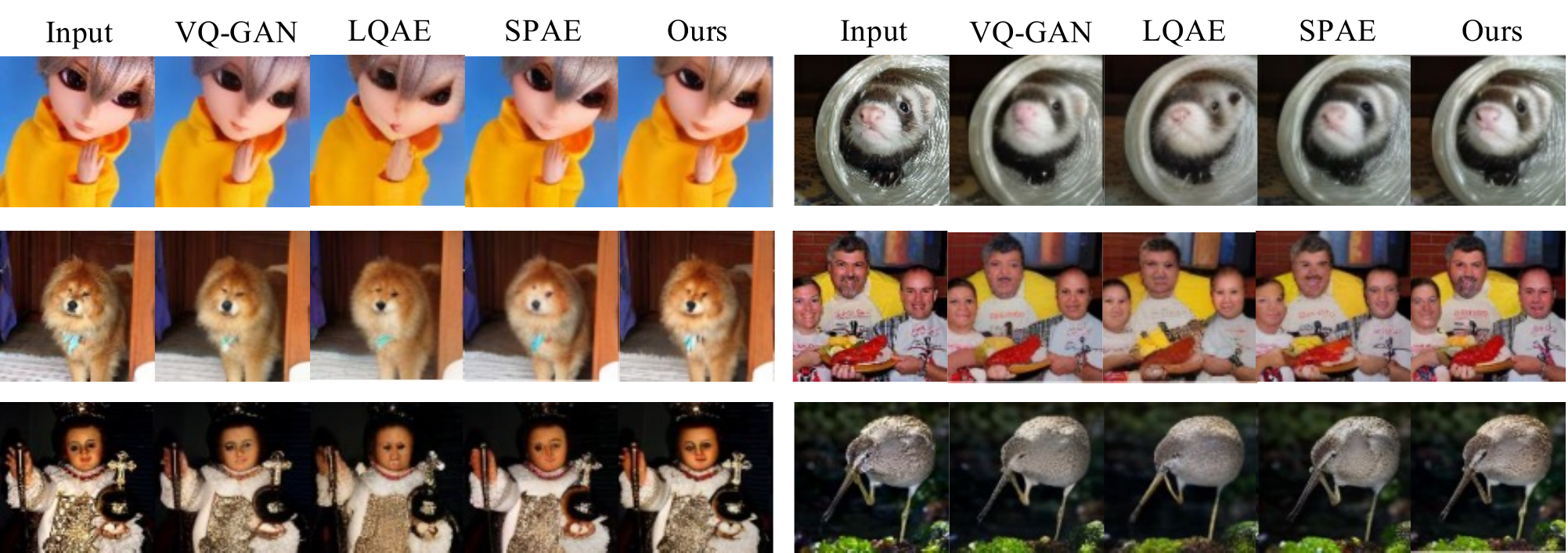}
\caption{Visualizations for image reconstruction.}
\label{fig:recons}
\end{figure*}

\noindent \textbf{Image Reconstruction.}
In Table~3 of the main paper, we report the quantitative results for reconstruction evaluation. In this study, we show several qualitative visualizations. In Figure~\ref{fig:recons}, we compare our approach with VQ-GAN~\cite{VQGAN}, LQAE~\cite{LQAE} and SPAE~\cite{SPAE}. Our approach is notable for its ability to reconstruct images with a high level of detail.

\begin{figure*}[t]
\centering
\includegraphics[width=0.99\textwidth]{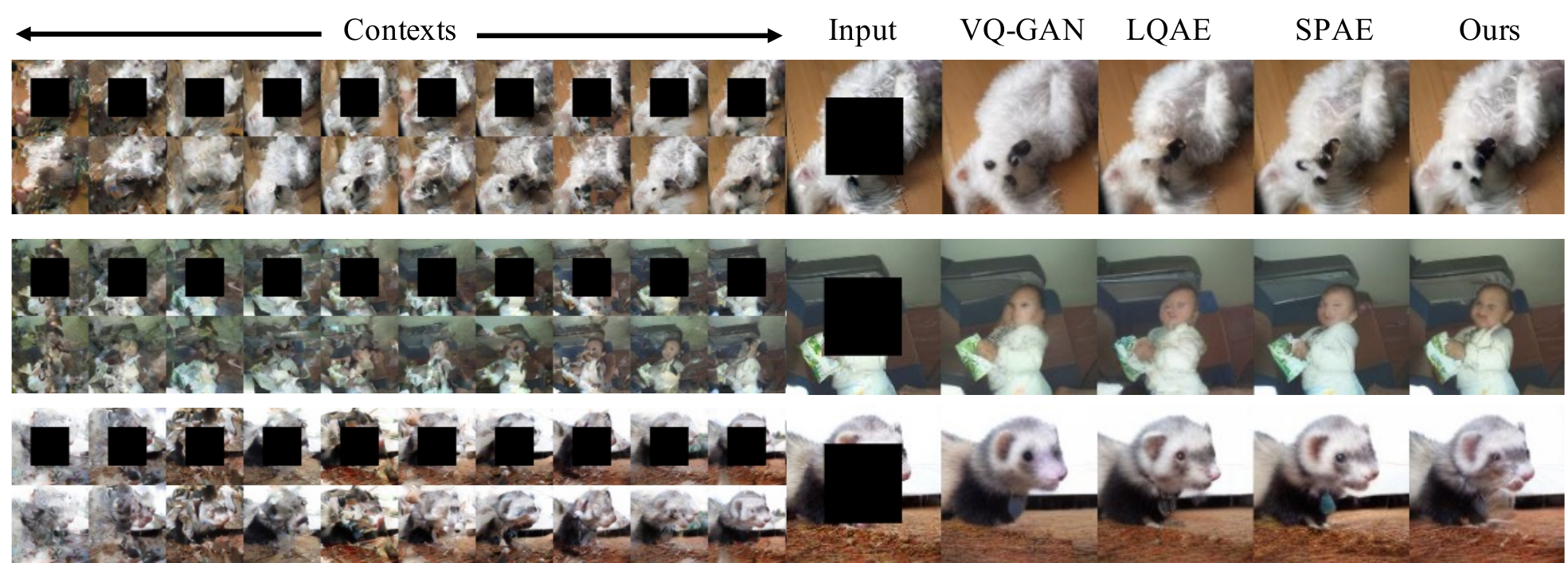}
\caption{Visualizations for image inpainting.}
\label{fig:inpainting}
\end{figure*}

\begin{figure*}[t]
\centering
\includegraphics[width=0.99\textwidth]{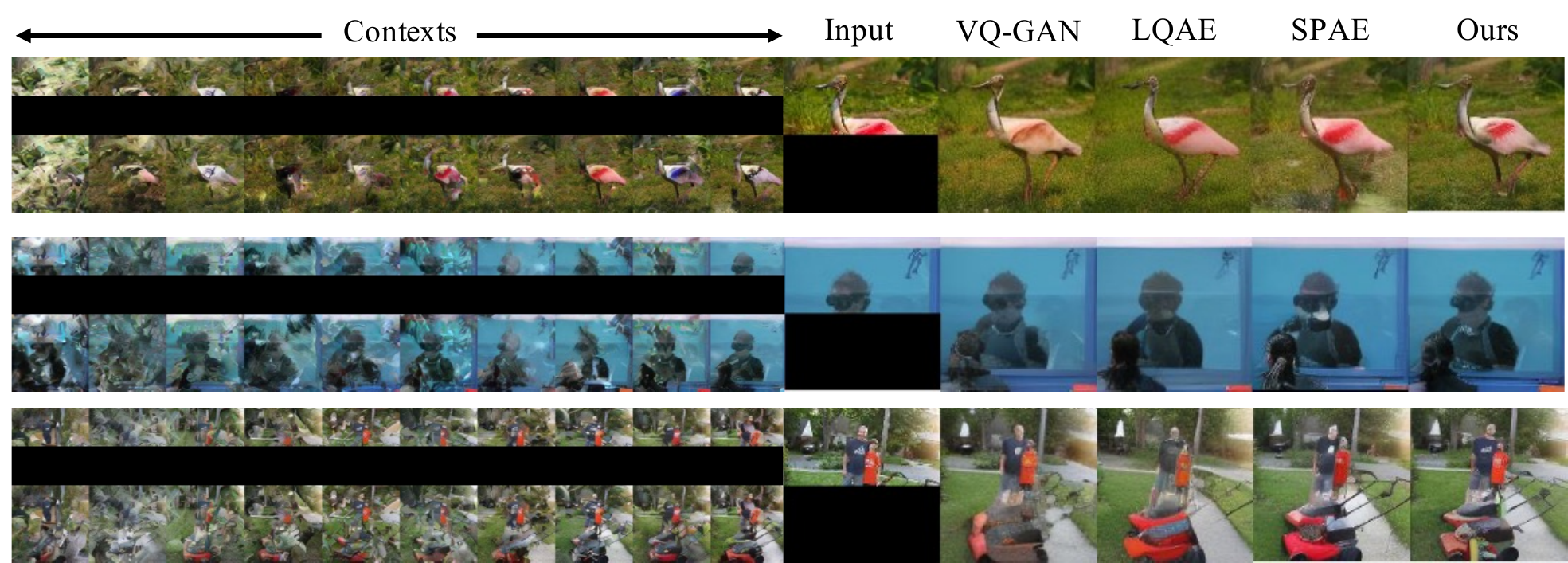}
\caption{Visualizations for image outpainting.}
\label{fig:outpainting}
\end{figure*}

\begin{figure*}[t]
\centering
\includegraphics[width=0.99\textwidth]{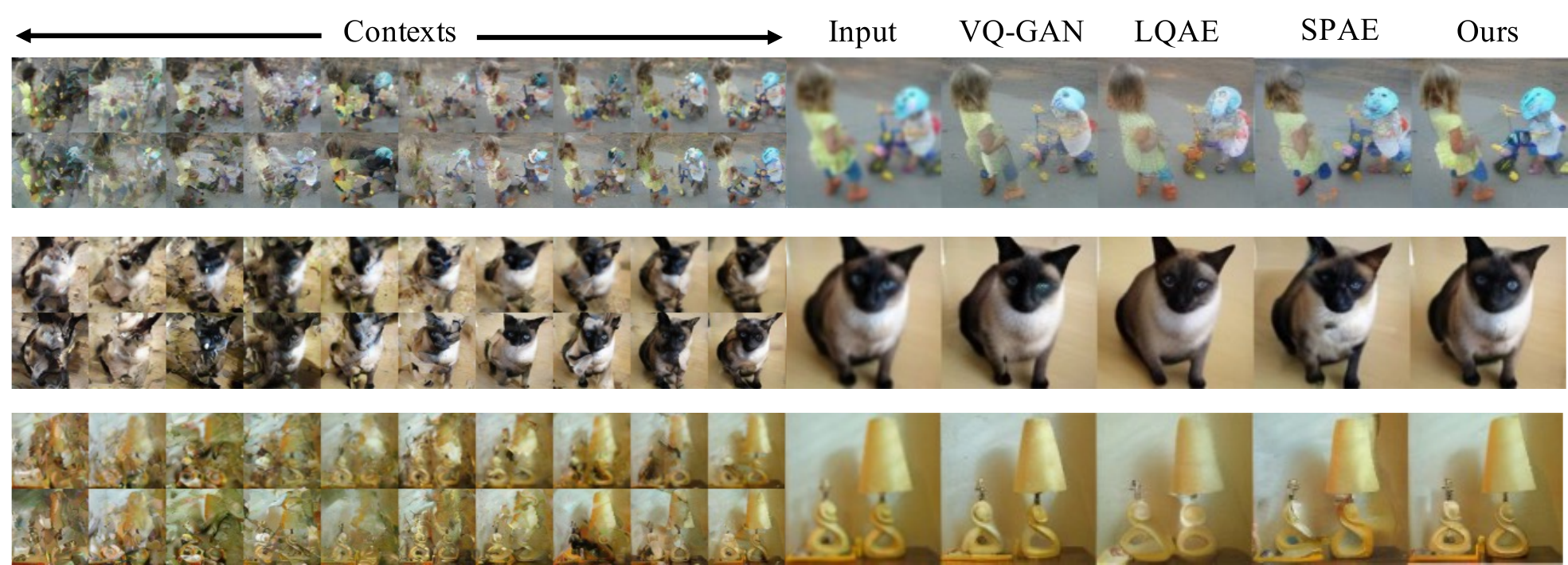}
\caption{Visualizations for image deblurring.}
\label{fig:deblur}
\end{figure*}

\begin{figure*}[t]
\centering
\includegraphics[width=0.99\textwidth]{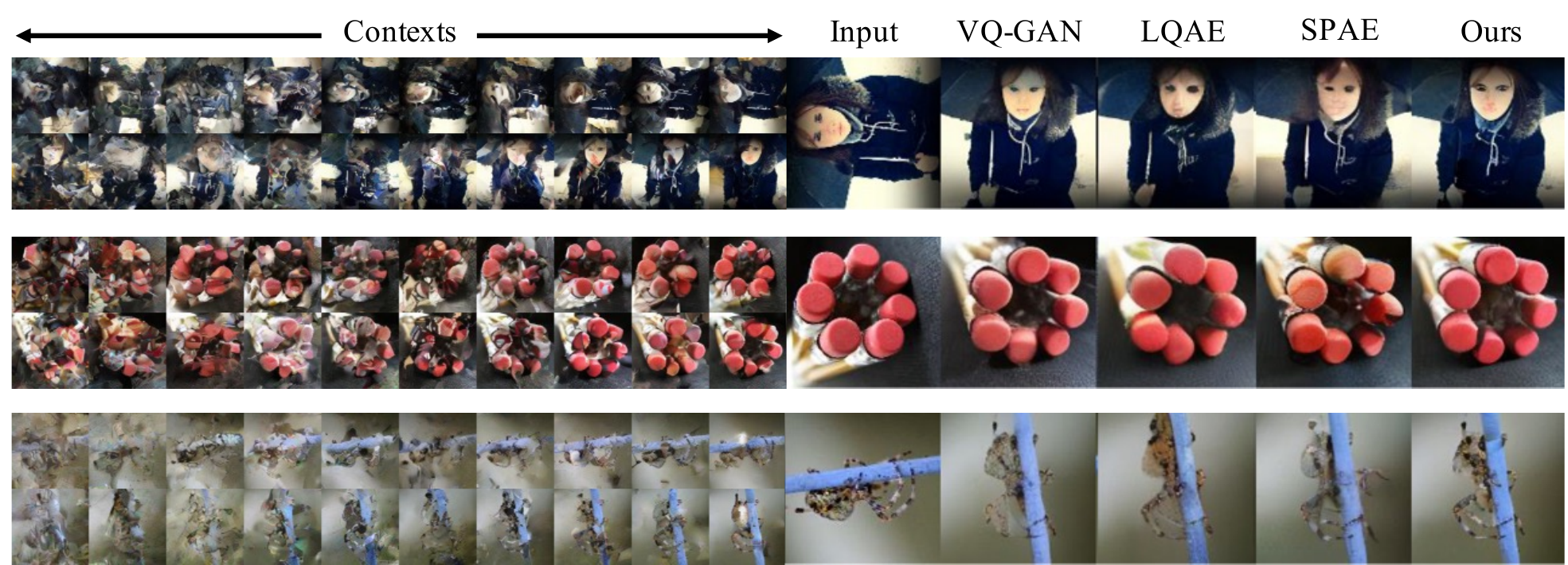}
\caption{Visualizations for rotation restoration.}
\label{fig:rotation}
\end{figure*}

\begin{figure*}[t]
\centering
\includegraphics[width=0.99\textwidth]{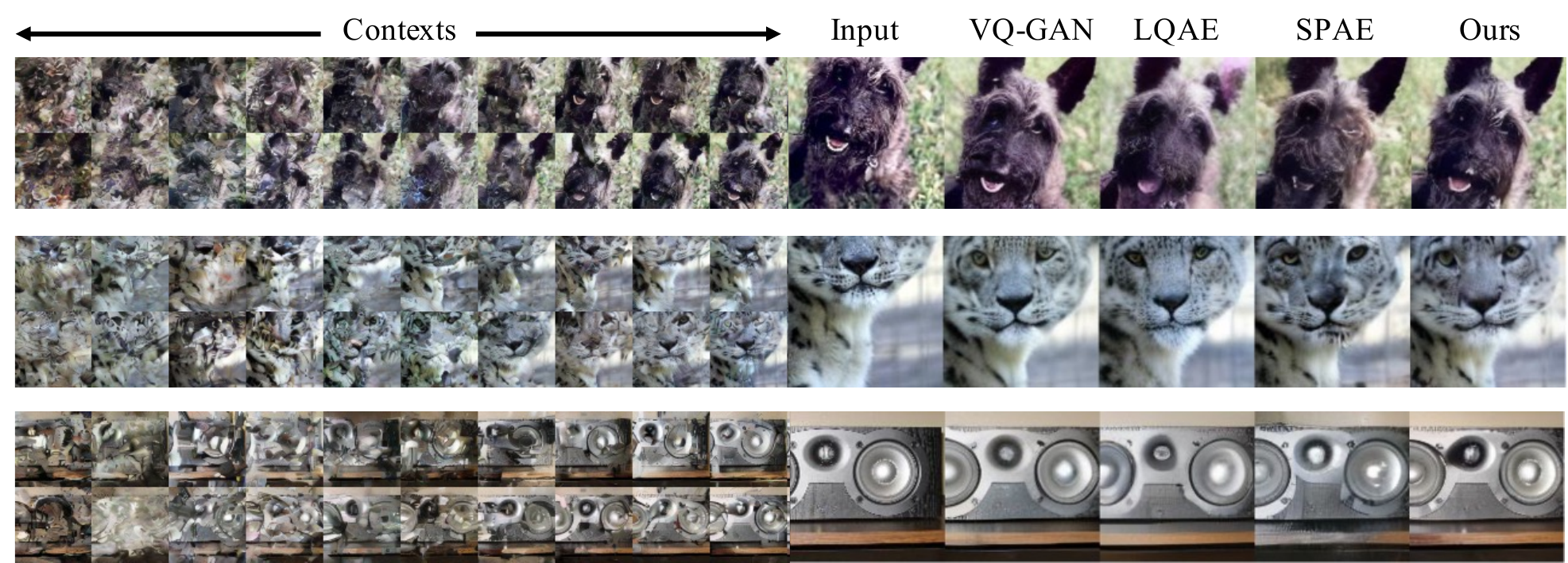}
\caption{Visualizations for shift restoration.}
\label{fig:shift}
\end{figure*}

\noindent \textbf{Image Denoising.} 
We show visualizations for image denoising in Figure 7 of the main paper. Here, we provide extra visualizations for inpainting (Figure~\ref{fig:inpainting}), outpainting (Figure~\ref{fig:outpainting}), deblurring (Figure~\ref{fig:deblur}), rotation restoration (Figure~\ref{fig:rotation}) and shift restoration (Figure~\ref{fig:shift}).

\end{document}